\documentclass[conference,compsoc]{IEEEtran}
\IEEEoverridecommandlockouts

\usepackage{cite}
\usepackage{amsmath,amssymb,amsfonts}
\usepackage{algorithm}                      %
\usepackage[noend]{algpseudocode}
\usepackage{graphicx,subfig}
\usepackage{booktabs}
\usepackage{multirow}
\usepackage{textcomp}
\usepackage{xcolor}
\usepackage[hyphens]{url}
\usepackage{flushend}

\usepackage{booktabs}
\usepackage[labelfont=bf]{caption}

\newlength{\tempheight}
\newlength{\tempwidth}
    
\newcommand{\rowname}[1]%
{\rotatebox{90}{\makebox[\tempheight][c]{\textbf{#1}}}}

\newcommand{\columnname}[1]%
{\makebox[\tempwidth][c]{\textbf{#1}}}

\newcommand{\comment}[1]{}    

\newif\ifcomment
\commenttrue
\ifcomment
	\newcommand\edc[1]{\textbf{\textcolor{blue}{EDC: #1}}	}
	\newcommand\mn[1]{\textbf{\textcolor{orange}{Mohammad: #1}}	}
	\newcommand\yf[1]{\textbf{\textcolor{cyan}{Yufei: #1}}	}
	\newcommand\edits[1]{{\textcolor{black}{#1}}	}
\else
	\newcommand\edc[1]{}
	\newcommand\mn[1]{}
	\newcommand\yf[1]{}
\fi

\newcommand{\descr}[1]{\smallskip\noindent\textbf{#1}}
    
\usepackage[square,sort,comma,numbers]{natbib}

\usepackage[bookmarks=true, bookmarksnumbered=true, colorlinks=true, linkcolor=linkcol, citecolor=citecol, urlcolor=urlcol, hypertexnames=true]{hyperref}

\definecolor{darkgreen}{RGB}{0, 100, 0}
\definecolor{linkcol}{rgb}{0.3,0,0}
\definecolor{citecol}{rgb}{0.3,0,0}
\definecolor{urlcol}{rgb}{0.3,0,0}
\definecolor{vlightgray}{gray}{0.925}

\usepackage[marginal,hang,stable]{footmisc}
\renewcommand{\footnoterule}{
	\kern -3pt
	\hrule width 1in
	\kern 2pt
}

\def\BibTeX{{\rm B\kern-.05em{\sc i\kern-.025em b}\kern-.08em
    T\kern-.1667em\lower.7ex\hbox{E}\kern-.125emX}}

\begin{document}

\title{BadVFL: Backdoor Attacks in Vertical Federated Learning$^*$
\thanks{$^*$Accepted for publication at the 45th IEEE Symposium on Security \& Privacy (S\&P 2024). Please cite accordingly.}

}

\author{\IEEEauthorblockN{Mohammad Naseri}
\IEEEauthorblockA{\textit{University College London}\\
mohammad.naseri.19@ucl.ac.uk}
\and
\IEEEauthorblockN{Yufei Han}
\IEEEauthorblockA{\textit{INRIA Rennes}\\
yufei.han@inria.fr}
\and
\IEEEauthorblockN{Emiliano De Cristofaro}
\IEEEauthorblockA{\textit{UC Riverside}\\
emilianodc@cs.ucr.edu}
}

\maketitle

\begin{abstract}
Federated learning (FL) enables multiple parties to collaboratively train a machine learning model without sharing their data; rather, they train their own model locally and send updates to a central server for aggregation. 
Depending on how the data is distributed among the participants, FL can be classified into Horizontal (HFL) and Vertical (VFL).
In VFL, the participants share the same set of training instances but only host a different and non-overlapping subset of the whole feature space. 
Whereas in HFL, each participant shares the same set of features while the training set is split into locally owned training data subsets. 

VFL is increasingly used in applications like financial fraud detection; nonetheless, very little work has analyzed its security.
In this paper, we focus on robustness in VFL, in particular, on backdoor attacks, whereby an adversary attempts to manipulate the aggregate model during the training process to trigger misclassifications.
Performing backdoor attacks in VFL is more challenging than in HFL, as the adversary i) does not have access to the labels during training and ii) cannot change the labels as she only has access to the feature embeddings. 
We present a first-of-its-kind clean-label backdoor attack in VFL, which consists of two phases: a label inference and a backdoor phase.
We demonstrate the effectiveness of the attack on three different datasets, investigate the factors involved in its success, and discuss countermeasures to mitigate its impact.
\end{abstract}


\section{Introduction}\label{sec:intro}
Federated learning (FL)~\cite{mcmahan2017communication} is increasingly used as a privacy-enhancing technique for distributed machine learning in a number of applications ranging from vision~\cite{Gong_2021_ICCV,Li_2021_CVPR} to fraud detection~\cite{DBA,creditrisk}. 
The main idea behind FL is to train local models on multiple datasets hosted separately by different participants.
Training does not explicitly require exchanging data samples; rather, each participant submits their local model to a central server, which aggregates them to generate a global model shared back to all participants.

FL can be classified into two kinds, depending on how the training data is split among participants.  
The most common setting is known as Horizontal FL (HFL), with the training data being split across the sample space.
That is, each participant hosts different training data points sharing the same feature set.
The participants also share the same classifier architecture. 
In each round of HFL, participants locally train, e.g., a classifier, and aggregate the local models to build a global classification model using aggregation algorithms like FedAvg or FedSGD~\cite{yang2019federated,mcmahan2017communication}.
Eventually, the model converges, and the parameters are finalized.

\descr{Vertical FL (VFL).} VFL adopts a split-learning architecture for training~\cite{VFL1, VFLUnleashing,VFL_Encrypted,VFL_split,fu2022label}.
Each local participant only hosts a subset of the (raw) features of the training data; more precisely, each of them maintains a bottom model to transform raw features to feature embedding vectors.
In each training round, the participants first share the feature embeddings with the server, which hosts the class labels of the training data.
The VFL server locally trains a top model for classification using the labels and the feature embeddings received from the participants.
Then, it propagates the gradient of the classification loss with respect to the feature embeddings so that each participant can update the bottom models using backpropagation.

As a rapidly emerging technology, VFL is particularly useful when the feature set to be analyzed is distributed across different organizations.
For example, bank account information and financial transaction records can be jointly used to evaluate the financial risk of an entity.
The two data sources are usually owned by different entities, e.g., banks and insurance companies.
By collaborating through VFL, organizations can gain valuable insights from data without compromising the confidentiality of sensitive information. %

\descr{Motivation.} Previous work has showed that HFL is prone to backdoor attacks~\cite{DBA,howtobackdoor,tails,alittle,ndss2021,durable}.
That is, adversarial participants can join the federated network and submit maliciously crafted models to mislead the joint model on trigger-embedded inputs.
Usually, backdoor attacks in FL work with trigger-free inputs and cause more harm to the integrity of FL systems than untargeted poisoning attacks, which deteriorate the accuracy of the global model indiscriminately on any input~\cite{chen2017targeted,xieICML21}. 
These attacks are often considered a roadblock preventing the trustworthy deployment of FL in security-critical applications, e.g., video surveillance~\cite{liu2020fedvision} or threat mitigation~\cite{MohammadCCS22}. 

While backdoor attacks in HFL have been studied extensively, it remains an open research problem to assess their feasibility and effectiveness in VFL.
In particular, \textit{VFL's split-learning design poses a unique challenge to successfully perform backdoor attacks like in HFL}.
First of all, participants in VFL cannot access or modify the labels hosted by the server. 
That is, it is hard for a participant acting as a backdoor attacker to produce backdoor training samples composed of pairs of trigger-embedded training features and the attacker-desired target labels~\cite{howtobackdoor}. 
As a result, we cannot simply adapt the attack approaches proposed in the context of HFL~\cite{DBA,howtobackdoor,tails,alittle,ndss2021,durable} to VFL. 

Second, each participant cannot access the feature subsets of other participants; the adversary can only manipulate the features that they control to perform poisoning. 
The flexibility of injecting backdoor perturbation to the feature space is thus limited, in contrast to backdoor attacks in HFL, where the attacker can change any of the features for the poisoning purpose.

\descr{Technical Roadmap.}  In this work, we present a first-of-its-kind study of backdoor attacks in VFL. 
First, as the attacker does not know the true labels of each training data instance, we set out to investigate whether backdoor attacks can be executed in VFL without access to the labels.
To do so, we design a novel backdoor attack, which we call \textit{BadVFL}.
The adversary relies on a label inference attack to reconstruct the labels of its data samples; then, she %
selects the source and target classes and injects a trigger using a saliency map (i.e., the most important regions) to some of her data samples, and submits the feature embeddings to the top model. 
This way, BadVFL aims to directly force the average distance between each backdoored data point and those of the target class to be close in the feature embedding space.

Second, we assess the effectiveness of the attack across different datasets and examine the factors contributing to its performance as well as its impact on the primary task accuracy.
Finally, we explore three possible countermeasures to mitigate BadVFL, namely, 1) backdoor defenses proposed in centralized learning like Neural Cleanse~\cite{wang2019neural}, 2) differentially private noise perturbation-based methods~\cite{ndss2021}, and 3) anomaly detection-based methods~\cite{DBA}. %

\descr{Main Findings.} Our work yields the following results:
\begin{enumerate}
	\item Clean-label backdoor attacks are feasible and pose realistic threats to model integrity in Vertical Federated Learning (VFL) systems when a fraction of the participants are adversarial. 
	For instance, in a two-party VFL setting using the CIFAR-10 dataset, the adversary can reach an Attack Success Rate (ASR) of 85\% with a random selection of source and target classes, while the main task accuracy is decreased by only 4\%. 
The attack can even be improved, through optimal selection of classes, to 89\% ASR, with a 5\% accuracy reduction in main task performance.\vspace{0.1cm}

	\item Different factors affect the effectiveness of the attack. 
	For instance, deciding when to abort the label inference phase and initiate the backdoor attack is a challenging task for the attack to be effective. 
	Moreover, the choice of the source and target classes impact both the attack's performance and the utility. 
	Other important factors that need to be considered as part of the adversarial strategy include trigger window size, poisoning budget, and auxiliary data.\vspace{0.1cm}
	\item The attack can be somewhat mitigated through differential privacy or anomaly detection, even though further work is needed to bring them to fruition. That is not the case for centralized learning defense methods, such as Neural Cleanse (NC)~\cite{wang2019neural}, as NC detects trigger signals in the input images to a machine learning system. Whereas in VFL, the server can only apply NC to the low-dimensional feature embeddings of the raw image, which makes NC less sensitive to the trigger-based perturbation in the feature space.
\end{enumerate}

\section{Preliminaries}\label{sec:preliminary}

\subsection{Vertical Federated Learning (VFL)}
In VFL, the sample space is the same for all participants' data, but the datasets differ in the feature space.
In other words, each participant hosts the same set of training instances but owns different and non-overlapping features of the same training instances.

VFL can have two architectures, VFL \textit{without}~\cite{VFL_Encrypted} and \textit{with} model splitting~\cite{VFL_split}.
The former follows a peer-to-peer decentralized learning process, requiring the participants to exchange intermediate results to compute the gradient/model updates.
We focus on the latter, i.e., VFL with model splitting, which follows the idea of split learning~\cite{VFL_split,fu2022label}.

In the model splitting setting~\cite{VFL_split,VFLUnleashing}, the model is divided into a top model and some bottom models.
The server hosts the top model as a classifier module. 
The bottom models, owned by the participants, are applied as encoders to transform the raw features of each participant into embedding feature vectors.
The top model receives the feature embeddings generated by the participants as input features, and the feature embeddings are then mapped to the corresponding class labels. 

We detail the training process for VFL with model splitting in Algorithm~\ref{alg:vflwithmodelsplitting}~\cite{fu2022label}.

\begin{algorithm}[t]
\small
\begin{algorithmic}[1]
\Require top model param $\theta_{top}$, bottom model params $\theta_1$, $\theta_2$, ..., $\theta_K$, true labels y, learning rate $\eta$
\Procedure{main}{}
\State Initialize $\theta_{top}$ and $\theta_1$, $\theta_2$, ..., $\theta_K$ 
\While{stopping epoch not met}
	\For{each batch $b$ of sample Ids}
		\For{$k=1$ to $K$}
			\State$o_k\gets $\Call{ParticipantForwardProp}{$\theta_k, b$}
		\EndFor
	\State$o_{all}\gets $\Call{ConcatBottomModels}{$o_1, ..., o_k$}
	\State$o_{final}\gets \theta_{top}(o_{all})$
	\State$L\gets LossFunc(o_{final}, y)$
	\State$g_{top}\gets \frac{\partial L}{\partial \theta_{top}}$
	\State$\theta_{top}\gets \theta_{top}-\eta \cdot g_{top}$
		\For{$k=1$ to $K$}
			\State$B_{k}\gets \frac{\partial L}{\partial o_k}$
			\State$B_k\gets B_k \cdot \frac{\partial o_k}{\partial \theta_k}$
			\State$\theta_{k}\gets \theta_{k} - \eta \cdot B_k$			
		\EndFor
	\EndFor
\EndWhile
\EndProcedure

\Function{ParticipantForwardProp}{$\theta, b$}
	\State return $\theta(b)$ \Comment{bottom model forward outputs}
\EndFunction
\end{algorithmic}
\caption{Training VFL with model splitting~\cite{fu2022label,VFL_split}}
\label{alg:vflwithmodelsplitting}
 \end{algorithm}

\subsection{VFL System Model}

We work with a VFL system involving $K$ participants ($K\geq{2}$) hosting separate feature subsets and a server. 
We denote the training features as $x_{i} \in R^{m}$ and label $y_{i}$ of one training instance $z_{i}$ as $x_{i} = \{x_{i,0},x_{i,1},x_{i,2},...x_{i,m}\}$.
The server owns only the class labels $\{y_{1},y_{2},y_{3},...,y_{n}\}$ of all $n$ training data instances.
   The $K$ local participants split the feature space in $R^{m}$ into disjointed subsets.
    Each of them privately owns one subset of the features of all the training instances.
These participants do not have direct access to the class labels, nor are they allowed to access the feature subsets of the others. 
We refer to them as {\em feature-hosting participants}. 

In a VFL system, for any given input $x_{i}$, each feature-hosting participant trains their bottom models (denoted as $B_{i}$), which transforms the hosted feature subspace $R^{m_{i}}$ ($m_{i}<m$) into a low-dimensional embedding space $R^{k_{i}}$, i.e., $E_{i}=B_{i}(\{x_{i,i_1},x_{i,i_2},...,x_{i,i_{m_i}}\})=\{e_{i,i_1},e_{i,i_2},...,e_{i,i_{k_{i}}}\}$.
The label-hosting server trains a top model as the classification module. It receives the feature embeddings $E_{i}$ from all the participants as input and maps them to the prediction of the class label.

The VFL training protocol is an iterative process. In each round of VFL training, each participant first submits the feature embeddings $E_i$ generated from their raw feature sets  (lines 5--6 in Algorithm~\ref{alg:vflwithmodelsplitting}). 
The label-hosting server receives the embeddings, aggregates and feeds them into the classification module (lines 7--8), updates the classification module, and computes the gradient of the loss function $\ell$ with respect to the embeddings of each participant (lines 10--11), $E_{i}$, $\partial \ell / \partial {E_{i}}$. The gradient vector is then propagated back to the local participants to update the parameters of their bottom models $B_{i} (i=1,2,3,...,K)$ (lines 13--15).

\section{The BadVFL Attack}\label{sec:attack}

In this section, we introduce the BadVFL attack.
First, we introduce the threat model, then summarize the challenges of building effective backdoor attacks in VFL.
Finally, we discuss BadVFL's attack methodology.

\subsection{Threat Model}\label{sec:threatmodel} %

\noindent{\bf Adversary's Goal.} We consider an adversary with a similar goal as traditional backdoor attacks against machine learning models~\cite{alittle}.
The adversary can operate in one of two modes:\smallskip

\noindent \textbf{\em 1) Single Attacker Mode}: One of the participants is controlled by the adversary, who aims to insert a trigger pattern $T$ into the features owned by the feature-hosting participants.
When the input instance embeds the trigger pattern $T$ into the corresponding feature subspace $R^{m_{i}}$, the classifier trained with VFL is supposed to predict an adversary-designated class label or perform normally otherwise.\smallskip

\noindent\textbf{\em 2) Multi-Attacker Mode}: We consider $M$ (for $M<K$) participants being controlled by the adversary.  
She decomposes a backdoor trigger $T$ into $M$ sub-trigger patterns and inserts each of the $M$ sub-trigger pattern  into the feature subspaces hosted by $M$ of $K$ feature-hosting participants. %
The mode aims to capture colluded backdoor attacks, whereby the compromised $M$ participants jointly enforce the backdoor learning task to memorize the trigger $T$ in the VFL classifier.  

\descr{Adversary's capabilities.} We assume that the adversary can access the training features hosted by the $M$ compromised participants. 
She can manipulate the feature subspaces hosted by the compromised feature-hosting participants, e.g., by inserting the backdoor trigger or sub-trigger patterns into the feature subspaces $R^{m_{i}}$ of each compromised feature-hosting participant $i$.
However, the adversary does not control the server; thus, she cannot directly access the labels of the training data nor the top model trained by the server.
Moreover, the adversary cannot access or manipulate the bottom models owned by the non-compromised feature-hosting participants. %

\descr{Adversary's knowledge.} The adversary can access the feature embedding $E_{i}$ committed by the bottom model of each compromised feature-hosting participant $i$. 
She can also access the gradient vectors $\partial \ell / \partial {E_{i}}$ sent by the server to the compromised participants. 
To mount an attack, we assume the adversary may collect a set of auxiliary data instances sharing the same feature distribution and label space as the true training data. 
This is a realistic assumption, as, e.g., the adversary can get additional images carrying the same labels beyond the true training data. 
This setting is also used in label inference attacks~\cite{fu2022label}.

\subsection{Challenges with Backdoor Attacks in VFL}\label{sec:challenges}
Despite progress in poisoning attacks against HFL systems, the difficulty of conducting backdoor attacks in VFL can be ascribed to the lack of access to: 1) the loss function and 2) the training labels. We now elaborate on this.

\descr{Loss function.} The malicious participants controlled by the adversary cannot access the loss function or the classifier used by the server.
Only the server can define the loss function to train the top-layer classification module or modify the architecture of the top-layer classifier function. 
Therefore, to launch a backdoor attack, the malicious participants cannot change the loss function of the classifier to adapt to the backdoor learning task as done in~\cite{Saha_Subramanya_Pirsiavash_2020}. 
Moreover, the malicious participants cannot introduce additional model architectures, such as GANs, to facilitate the attacks~\cite{Nguyen2020NIPS}, but must follow the server's learning loss in the attack process. 

\descr{Training labels.} The malicious participants cannot directly access or modify the training labels used by the server.
Therefore, unlike in classical backdoor attacks, the adversary cannot directly introduce the backdoor training samples composed of pairs of the trigger-embedded feature vectors and the attack-desired class label into the VFL training set. 

\subsection{Attack Methodology}\label{sec:algorithm}
To address these challenges, our novel attack, which we call BadVFL, follows a two-staged attack pipeline. 
We first describe it for the {\em Single Attacker Mode}. %

\descr{Model Extraction Stage.} Given a feature-hosting participant $P_{k}$ in the VFL architecture compromised by the adversary, $P_{k}$ first infers the labels of the training samples $\{x_{i}\}$. Each $x_{i}$ on this malicious participant is composed of the values of the feature subset assigned to the participant, denoted as  $x_{i,i_{1}},x_{i,i_{2}},x_{i,i_{3}},...,x_{i,i_{m}}$. 
It is easy for the adversary to collect a small set of labeled / partially labeled auxiliary data instances sharing the same label space. 
The auxiliary data samples do not overlap with those in the true training data. 
We denote the auxiliary training data set as $S$. 
The adversary can feed these auxiliary training instances to the bottom model of the malicious participant (noted as $B_{\texttt{adv}}$) hosted by $P_k$ to produce the feature embeddings of the auxiliary data.

Next, the adversary can train a classifier using the feature embeddings and labels of the auxiliary data via a standard supervised learning protocol or a semi-supervised learning method if the auxiliary data instances are partially labeled. 
The classifier will be used as a surrogate model to the top model trained by the server; we denote it as $\hat{h}$. 
$\hat{h}$ is assumed to have a different model architecture from the true top model; in reality, the malicious participant should not know the definition of the top model or how it is trained.

The result of the model extraction stage is to locate the source and target classes of the BadVFL attack, as well as identify the training data instances of the two classes. 

\descr{Backdoor Trigger Insertion Stage.}  We denote the source and target classes of training data instances for the backdoor attack as $\{x^{s}_{i}\} \in \mathcal{D}_{s}$ and $\{x^{t}_{i}\} \in \mathcal{D}_{t}$, respectively. 
We randomly pick a subset $\mathcal{D}^{\text{sub}}_{s}$ from the source class of data $\mathcal{D}_s$.
Next, we follow the strategy of clean-label backdoor attacks~\cite{Zhao2020cvpr} to inject the trigger signal to the training samples in $\mathcal{D}^{\text{sub}}_{s}$. 
The objective function of inserting the trigger into the VFL system is given in Equation~\ref{eq:backdoor} below.
 
The goal is to force the feature embeddings of the perturbed training samples from $\mathcal{D}^{\text{sub}}_{s}$ embedded with the adversary-designated trigger signal in the source class to move close to those of the training instances of the target class. 
The dirty-label backdoor attacks\cite{howtobackdoor} directly introduce the pairs of trigger-embedded training features and adversary-desired class labels into the training data to set up the association between the trigger-embedded input and target class label. 
Nevertheless, the adversary in VFL cannot change the class labels of the training data set, as these are privately owned by the server. 
Thus, an alternative way is to perturb some training instances of the target class to drag them close to the trigger-embedded instances in the source class. 
The intuition is that if the perturbed instances of the target class are similar enough to the trigger-embedded instances of the source class, the trained classifier will not be able to differentiate them and thus misclassify the trigger-embedded instances as the target class. 

Equation~\ref{eq:backdoor} is as follows:
\begin{equation}\label{eq:backdoor}
\small
\begin{split}
B^{*}_{\texttt{adv}} &= \underset{B_{\texttt{adv}}}{\arg\min}\,\, \sum_{i}\|B_{\texttt{adv}}(\hat{x}^{\text{sub},s}_{i} ) - B_{\texttt{adv}}(x^{t}_{i})\|^2_{fro} \\
s.t.\,\,\,\,& \hat{x}^{\text{sub},s}_{i} = {x}^{\text{sub},s}_{i} + \delta, \,\,\,\, \|\delta\|_{L2} \leq {\epsilon}\\
\end{split}
\end{equation}
where $\delta$ is the backdoor trigger specified by the adversary and embedded in the training data instances of the source class. 
${x}^{\text{sub},s}_{i}\in \mathcal{D}^{\text{sub}}_{s}$ is the training samples selected from the source class to inject the backdoor trigger.  
$B_{\texttt{adv}}({\hat{x}^{\text{sub},s}_{i}})$ and $B_{\texttt{adv}}(x^{t}_{i})$ are the feature embeddings of the trigger-perturbed samples in the source class and those of the perturbation-free samples in the target class respectively. We include 
$\|\|_{fro}$ and $\|\|_{L2}$ are the Frobenius and L2 norm. $\epsilon$ is a designated bound of the poisoning noise injected into the target class of training data. 

In addition to setting up the mapping between the triggered instances in the source class and the target class label in the embedding space, we also require the backdoor perturbation injected to the selected data $\mathcal{D}^{\text{sub}}_{s}$ to be as small as possible, as posed by the L2 distance based constraint in Equation~\ref{eq:backdoor}. 

\begin{figure}[t]
\centering
\includegraphics[width=0.45\textwidth]{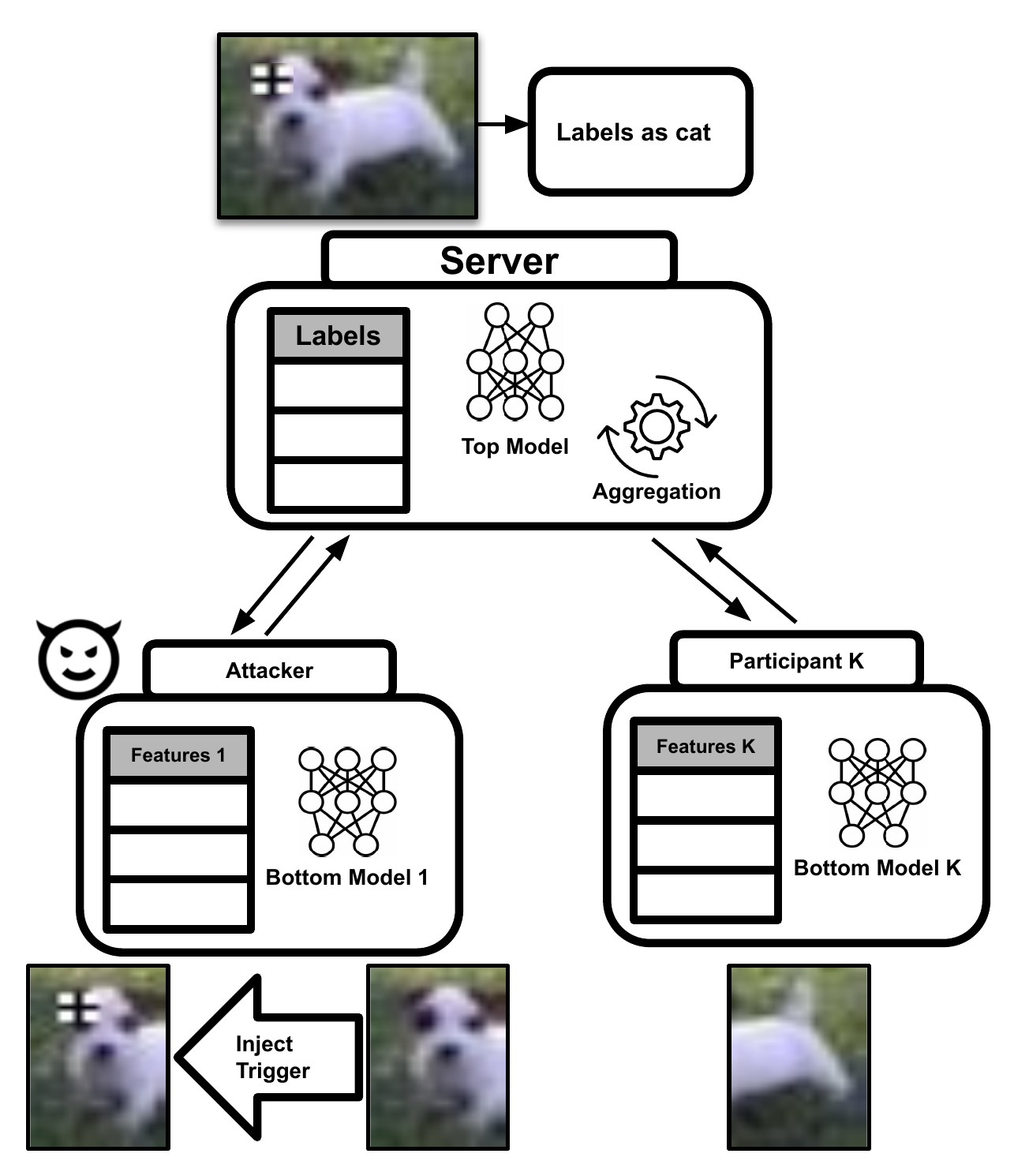}
\caption{An example of how the BadVFL attack works on a data point in a two-party VFL setting with one attacker.}
\label{fig:attack_example}
\end{figure}

We aim to minimize trigger-induced modifications to the backdoor perturbed training data $\mathcal{D}^{\text{sub}}_{s}$ in the source class to make sure that the classifier can misclassify the trigger-embedded input to the target class. Too much perturbation may completely change the contents of the perturbed data $\mathcal{D}^{\text{sub}}_{s}$ and acts as a smoking gun to human annotators. Furthermore, arbitrarily large trigger perturbation can cause uncontrollable bias to the feature embeddings of the poisoned data, which fails attack. Note that this constraint is widely adopted in poisoning attacks of machine learning models \cite{geiping2021witches}, which is employed to avoid introducing unexpectedly large poisoning noise to the training data and smoothing the learning process of data poisoning.

Learning the perturbation noise is, in general, a challenging non-convex and high-dimensional optimization problem. 
By anchoring the perturbed training data around the noise-free training data, learning the perturbation noise using gradient descent can be made stable and effective. 
Once we generate the backdoored training data $\{\hat{x}^{sub,s}_{i}\}$ belonging to the source class s, we replace the selected subset $\mathcal{D}^{\text{sub}}_{s}$ with the corresponding poisoned data $\mathcal{\hat{D}}^{\text{sub}}_{s}$. 
With this perturbed training dataset, we resume tuning the bottom model $h$ following the training protocol of VFL. 
An example of backdoogr injection in a two-party VFL setting by a single attacker is illustrated in Figure~\ref{fig:attack_example}.

\descr{Multi-Attacker Mode.} Finally, we extend BadVFL to work with more than one adversary (thus, the number of participants must be greater than two).
In this setting, the attacker is required to divide the trigger into sub-triggers. These sub-triggers are then assigned among the participants she controls, which makes the attack more challenging.
Also, the auxiliary dataset is now shared among all compromised participants.
This means that each attacker has access to the same set of auxiliary data, which may help their attacks. 

The malicious participants start by conducting label inference attacks individually but vote to achieve more accurate label estimation. When they have reached a consensus on the label estimate, they introduce triggers to the source and target classes simultaneously.

\section{BadVFL Pipeline}\label{backdooralgorithmflow} 
In this section, we discuss in detail the four-step pipeline of the BadVFL attack.
We start with the single-attacker mode; the pipeline can be applied to the multi-attacker mode with minor changes.  
Without loss of generality, we assume two participants in a VFL system, one of them controlled by the adversary. 
The other feature-hosting participant and label-hosting server are both benign. 

\descr{Step 1:} The adversary estimates the labels of the training instances hosted by the malicious participant using the labeled auxiliary dataset $S$.%
 
\descr{Step 2.1:}  The adversary randomly picks two classes of training images, i.e., class $S$ (the source class) and class $T$ (the target class), denoted as \textit{Random Selection}. 
The adversary can clone the training images of the source class and inject the backdoor trigger into the cloned training images.
We denote the training images of class $S$ and $T$ as $\{x^{S}_{i}\})$ (i=1,2,3,...,$N_{S}$) and $\{x^{T}_{j}\}$ (j=1,2,3,...,$N_{T}$).
The cloned and backdoored images of class $S$ are denoted as $\{\hat{x}^{S}_{i}\} (i=1,2,3,\ldots,N_{S})$, where $\hat{x}^{S}_{i} = x^{S}_{i} + \delta$, with $\delta$ being the backdoor trigger.

Besides \textit{Random Selection}, the adversary can also choose the source and target classes for the backdoor attack based on the averaged pairwise distance between the feature embeddings of the source and target class. 
We name this setting as \textit{Optimal Selection} in the following. 
Essentially, after the label inference phase, the adversary roughly knows the labels of each data point.
In \textit{Optimal Selection}, she computes the average pairwise distance between the training data points in the two classes. 
The value denotes the distance between two classes of data points in the embedding space.
The adversary repeats the process for every pair of classes and eventually finds the pair of classes with the least average pairwise distance. 
Intuitively, if the feature embeddings belonging to the source and target class are close to each other, it will be easy for the adversary to make the poisoned training samples in the target class similar to the trigger-embedded training instances of the source class in the embedding space.
This can maintain the same level of attack performance while simultaneously improving the overall performance of the main task.
We will discuss how \textit{Optimal Selection} improves the backdoor attack later in our experimental evaluation. 

\descr{Step 2.2:} The adversary then computes the Jacobian-based saliency map (gradient map) \cite{papernot2017blackbox} of the training images $\{x^{S}_{i}\}$ using the locally trained classifier $\hat{f}$.
We use $G_{i}$ to denote the saliency map of a training image $i$ hosted by the adversary.
The adversary uses a 3-by-3 (or 5-by-5) sliding window to deliver a pass over a source training image in $\{\hat{x}^{S}_{i}\}$.
She chooses the sliding window with the highest average gradient magnitudes to inject the backdoor trigger $\delta$. 
The saliency map measures the sensitivity of the classification loss with respect to each area of the input image. 
High/low saliency values imply that any modification to the corresponding image area causes large/small fluctuations in classification loss. 
We choose the image areas of high saliency values as the target to inject backdoor triggers. 
Our aim is to bring as much change as possible to the classification loss by adding the trigger of the limited size. 

\descr{Step 3:} The adversary randomly picks $p\%$ of samples from $\{x^{T}_{i}\}$ and replaces them with the corresponding poisoned training data learned by Equation~\ref{eq:backdoor}.
The adversary tunes the poisoning noise added to the selected training data from the target class according to Equation~\ref{eq:backdoor}. 
The modified set of training instances is denoted as $\hat{D}$.
The feature embeddings of $\hat{D}$ (noted as $B_{\texttt{adv}}(\hat{D})$).
The attack strength is proportional to the number of backdoored training instances; i.e., more injected backdoor training samples (a larger $p\%$) yield stronger attacks and vice versa.

\descr{Step 4:} The adversary receives the gradient of the loss function with respect to the submitted embedding $B_{\texttt{adv}}({\hat{D}})$ and propagates back the gradient to different parameters of the bottom model $B_{\texttt{adv}}$. 
The VFL training process is conducted using Step 3 and Step 4 iteratively until the training process converges.

\descr{Multi-Attacker Mode.} In this mode, we assume there are $K>2$ participants in the VFL system, and at least two of them are malicious. %
The malicious participants vote to achieve consensus over the label estimates of the training data. 
Based on that, each conducts the backdoor trigger injection step individually using the assigned sub-triggers following the steps from Step 2.2 to Step 4. During the injection stage, the malicious participants use the same poisoning budget $p\%$. 

\section{Experimental Evaluation}\label{sec:exp}
\edits{
This section measures BadVFL's effectiveness, starting with a single attacker and two VFL participants.
We investigate the impact of several factors in the attack setting, i.e., the number of VFL training rounds used for the model extraction stage, how the source/target classes are selected, the poisoning budget, how the trigger is inserted into the training data, and the size of the backdoor trigger for image datasets. 
}
\edits{
We then extend our evaluation to the multi-party setting ($K>2$ participants), considering both single and multi-attacker modes. 
}

\subsection{Dataset}
\edits{
Our experiments use three image datasets (CIFAR-10~\cite{krizhevsky2009learning}, CIFAR-100~\cite{krizhevsky2009learning}, and CINIC-10~\cite{darlow2018cinic}) and one tabular dataset (Criteo~\cite{criteo}).
In CIFAR-10, the training and testing set sizes are, respectively, 50,000 and 10,000, and the number of classes is 10.
CIFAR-100 has the same number of images but 100 classes. 
CINIC-10 has 10 classes and 270,000 images, of which 180,000 are used for training and 90,000 for testing.
In all image datasets, images are 32$\times$32 pixels. 
Criteo is a real-world dataset used for predicting ad click-through rates, which employs both categorical and continuous features.
We choose it as embedding backdoor signals into Criteo features, leading to incorrect click-through rate estimations, might cause commercial leaders to make wrong selling/purchasing decisions with respect to one or several specific products.
The dataset consists of 80,000 training and 20,000 testing samples, and the number of classes is 2. 
}

\edits{
We use the same training and testing sizes and follow the model splitting setting proposed in~\cite{fu2022label}. 
For the image datasets, we use a ResNet-18 architecture for the bottom model, while, for the top model, the feature embeddings generated by ResNet-18 are taken as input to a Fully Connected Neural Network 4 (FCNN-4).
For the Criteo dataset, we use a Fully Connected Neural Network 3 (FCNN-3) for both bottom and top models. 
}

\subsection{Attack Settings}
We focus on both two-party and multi-party VFL settings. 
Unlike HFL, which can often include thousands of participants, VFL typically involves only a few; in fact, two-party VFL seems to be the most common setting~\cite{chen2021homomorphic,fu2022label}. 

\edits{
For the three image datasets, we split them vertically into $K$ parts ($K$ being the number of participants in the VFL system), as done in previous work~\cite{liu2019communication,liu2020asymmetrical,fu2022label}. 
For Criteo, we adhere to the method outlined in~\cite{fu2022label}.
Although this includes categorical and numerical attributes, we only consider the latter and ignore the former for the attack; for this reason, all numerical attributes are initially projected into a hash space of $2^{13}$ dimensions. 
We then halve the dimensions of the attributes, resulting in both the benign participant and the adversary holding $2^{12}$ dimensions each.}

\edits{For image datasets, we use 5$\times$5 and 3$\times$3 pixel sliding windows separately to inject the trigger into a target image.
The attacker selects half of the numerical features and changes their value to a fixed value that falls outside the usual range of values.
Following~\cite{yang2022clean}, we gradually increase the percentage of poisoning data points (denoted as $p\%$ in the algorithm flow) to traverse different \textit{poisoning budget} values in our experiments.}
 
To measure the effectiveness of the backdoor attack, we measure the \textit{Attack Success Rate} (ASR) of BadVFL over the testing data. 
More precisely, ASR is computed as the fraction of the backdoor samples that are correctly classified as the attacker-desired class.

Recall that BadVFL consists of two phases: model extraction and backdoor poisoning (see Section~\ref{backdooralgorithmflow}).
To execute the attack, the total number of rounds should be divided between the two phases. 
We denote the total number of VFL training rounds with $\texttt{TotalRounds}$. 
We assume up to $R_n$ rounds are used for extracting the classification module and estimating the labels.  
The remaining $\texttt{TotalRounds}-R_n$ rounds are dedicated to generating poisoning samples in the feature embedding space and performing backdoor attacks following the VFL protocol. 

\edits{
We evaluate the performance of the attack using both \textit{Random Selection} and \textit{Optimal Selection} of the source and target classes.
For the former, we assume that the adversary randomly picks two classes; for the latter, the adversary uses the estimated class labels to categorize training data hosted by the malicious participants into separate groups. 
She then computes the average pairwise distance between a pair of groups in the feature embedding space of the bottom model owned by the adversary. 
The distance measures how close the two groups of training samples are in the embedding space. 
Finally, the adversary chooses the two classes with the least average pairwise distance as the source and target class of the backdoor attack. 
}

\begin{figure}[t]
\centering
\includegraphics[width=0.35\textwidth]{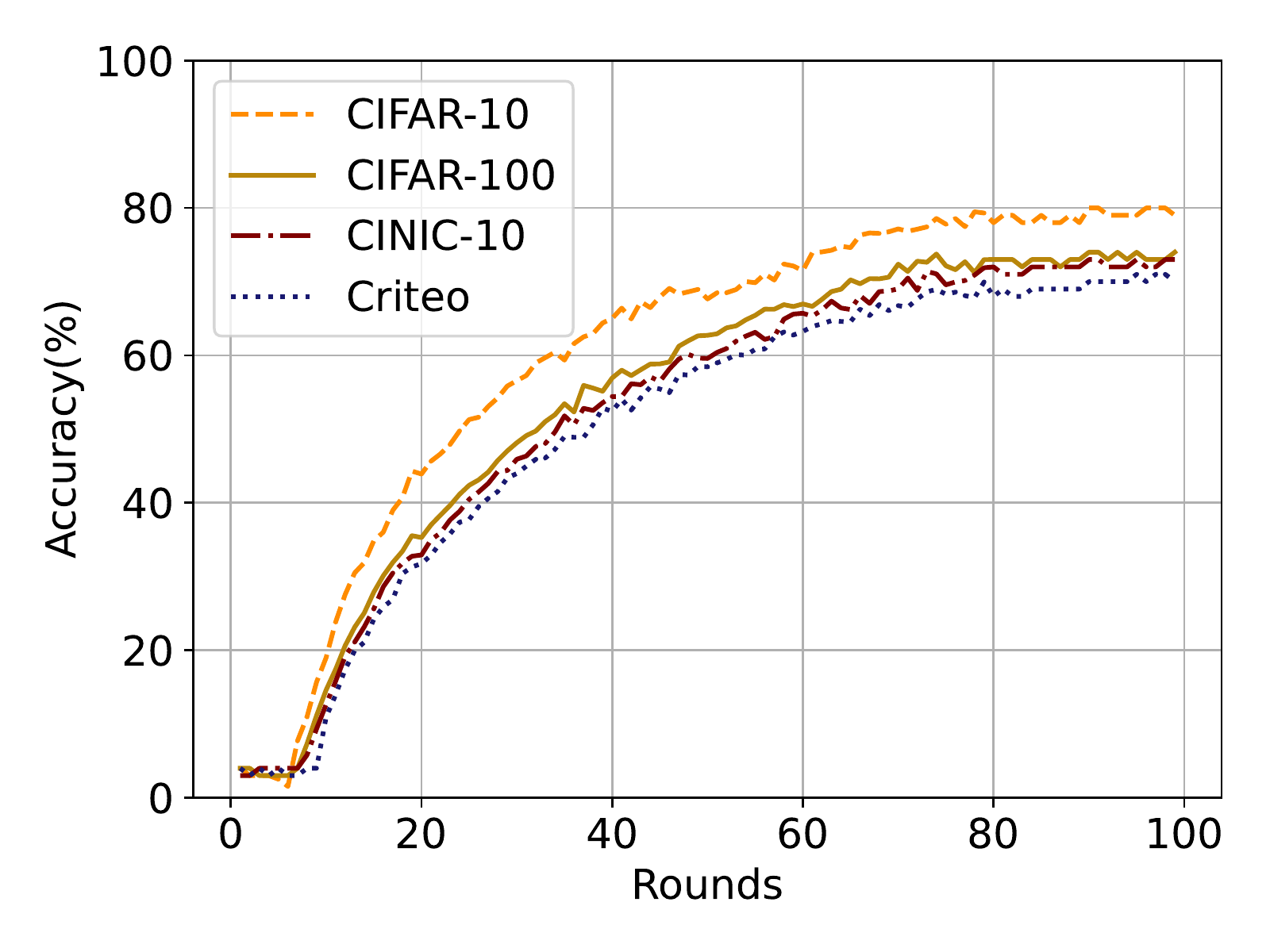}
\caption{\edits{Main Task Accuracy (MTA) over 100 rounds with no attack.}}
\label{fig:two-party-acc-by-round-no-attack}
\end{figure}

\begin{figure*}
\centering
\subfloat[CIFAR-10]{\label{CIFAR-10backdoored3x3}\includegraphics[width=0.13\textwidth]{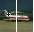}}\hspace*{-0.01cm}
\subfloat[CIFAR-10]{\label{CIFAR-10backdoored5x5}\includegraphics[width=0.13\textwidth]{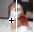}}\hspace*{-0.01cm}
\subfloat[CIFAR-100]{\label{CIFAR-100backdoored3x3}\includegraphics[width=0.13\textwidth]{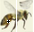}}\hspace*{-0.01cm}
\subfloat[CIFAR-100]{\label{CIFAR-100backdoored5x5}\includegraphics[width=0.13\textwidth]{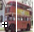}}\hspace*{-0.01cm}
\subfloat[CINIC-10]{\label{CINIC-10backdoored3x3}\includegraphics[width=0.13\textwidth]{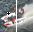}}\hspace*{-0.01cm}
\subfloat[CINIC-10]{\label{CINIC-10backdoored5x5}\includegraphics[width=0.13\textwidth]{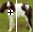}}\hspace*{-0.01cm}
\caption{Examples of images (left side owned by the attacker) embedded with the backdoor trigger in three datasets. Two sliding window sizes, 3$\times$3 and 5$\times$5, are used for this purpose.}
\label{backdooredimages}
\end{figure*}

\subsection{Two-Party Experiments}
\label{twopartysection}
\edits{
In the two-party experiments, the features are divided among two participants.
For image datasets, each data sample (an image) is vertically split into two halves from the middle, with each participant holding half.
Following~\cite{fu2022label}, we assume the adversary has an auxiliary dataset of, respectively, 40, 400, 40, and 100 data points, uniformly sampled from each class of the training datasets of, respectively, CIFAR-10, CIFAR-100, CINIC-10, and Criteo. 
}

Fig.~\ref{fig:two-party-acc-by-round-no-attack} reports the accuracy of the main learning task during 100 rounds. %
We set the total number of rounds ($\texttt{TotalRounds}$) to 100 for an aggregated model with acceptable classification performance over the testing set. 
Table~\ref{maintaskperformance_twoparty} reports the accuracy of the main task model, along with precision and recall, under no attack after 100 rounds. 
The results are averaged over five runs of sampling training/testing data using a with-replacement sampling strategy.

\begin{table}[t]
\centering
\begin{tabular}{lrrr}
\toprule
\textbf{Dataset}& \textbf{Accuracy}& \textbf{Precision}& \textbf{Recall}  \\  
\midrule
CIFAR-10 & 0.81   & 0.83   & 0.78       \\ 
CIFAR-100 & 0.74 & 0.75 & 0.70     \\ 
CINIC-10& 0.73 & 0.72 & 0.72  	  \\ 
Criteo & 0.71 & 0.74 & 0.69  	  \\ 
\bottomrule
\end{tabular}
\caption{Performance of the main task in the two-party setting, under no attack, after 100 rounds.}
\label{maintaskperformance_twoparty}
\end{table}

\begin{figure*}
\centering
\subfloat[CIFAR-10]{\label{3metrics100rounds_2party_cifar10}\includegraphics[width=0.255\textwidth]{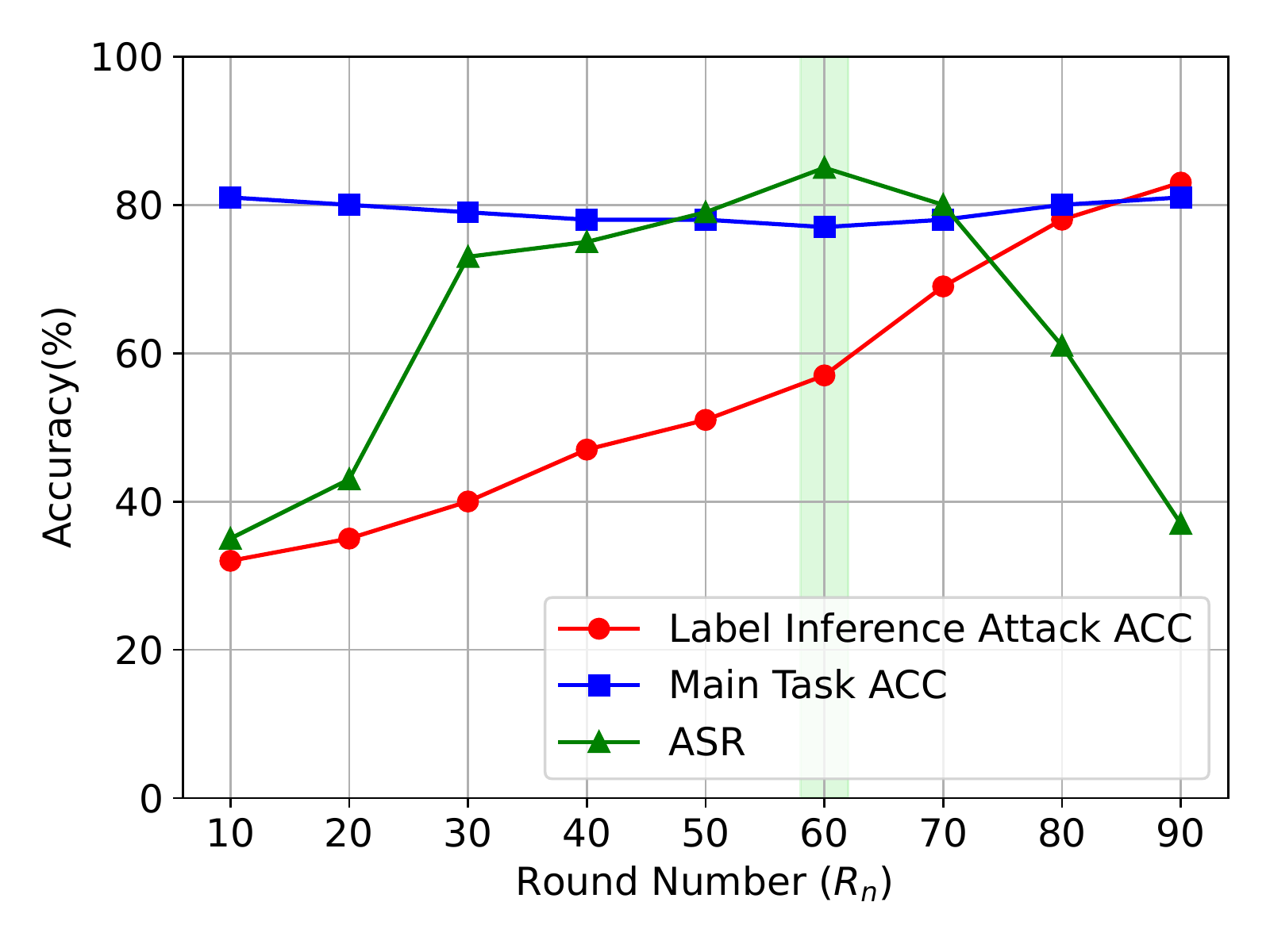}}\hspace*{-0.2cm}
\subfloat[CIFAR-100]{\label{3metrics100rounds_2party_cifar100}\includegraphics[width=0.255\textwidth]{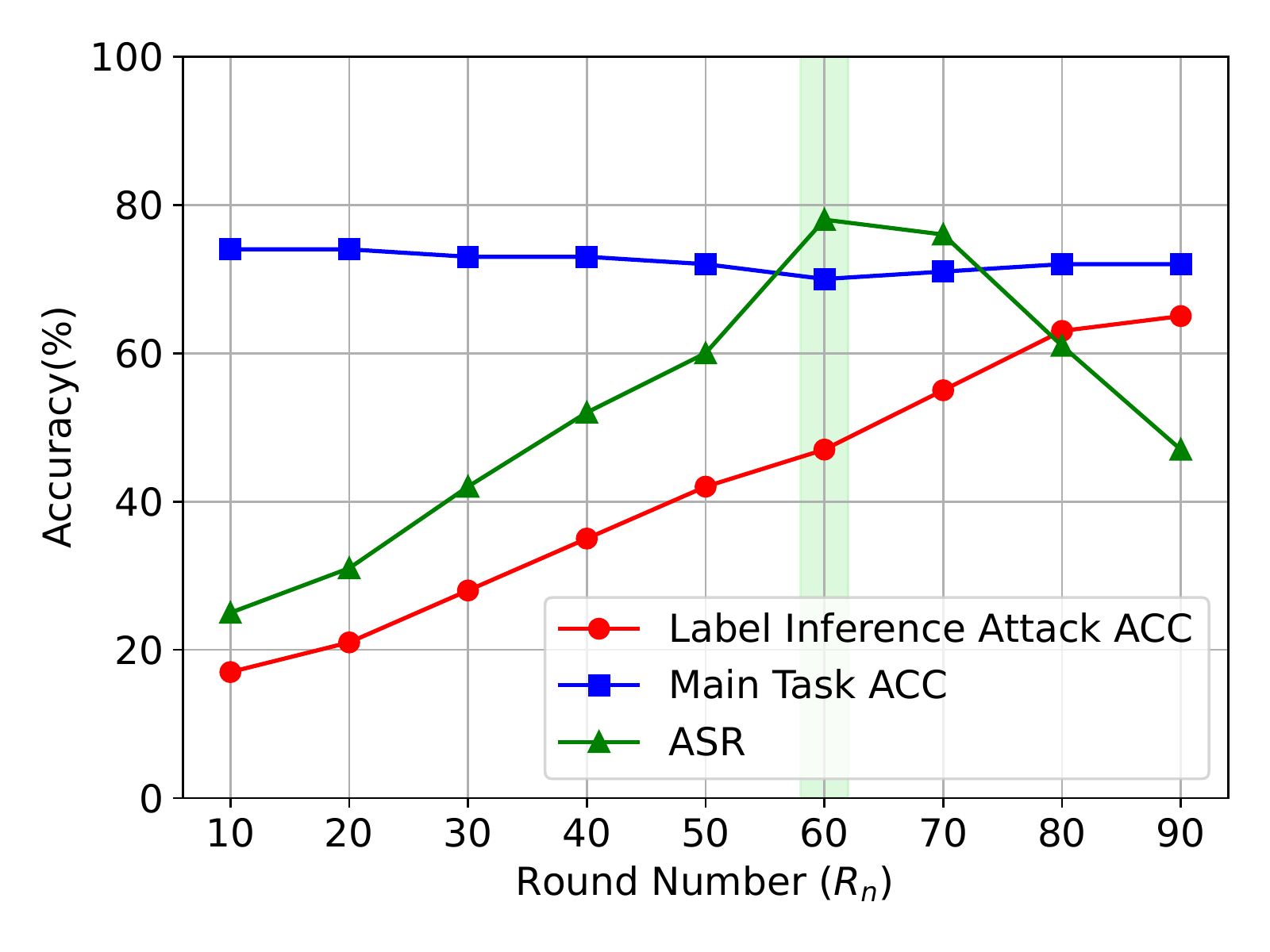}}\hspace*{-0.2cm}
\subfloat[CINIC-10]{\label{3metrics100rounds_2party_cinic10}\includegraphics[width=0.255\textwidth]{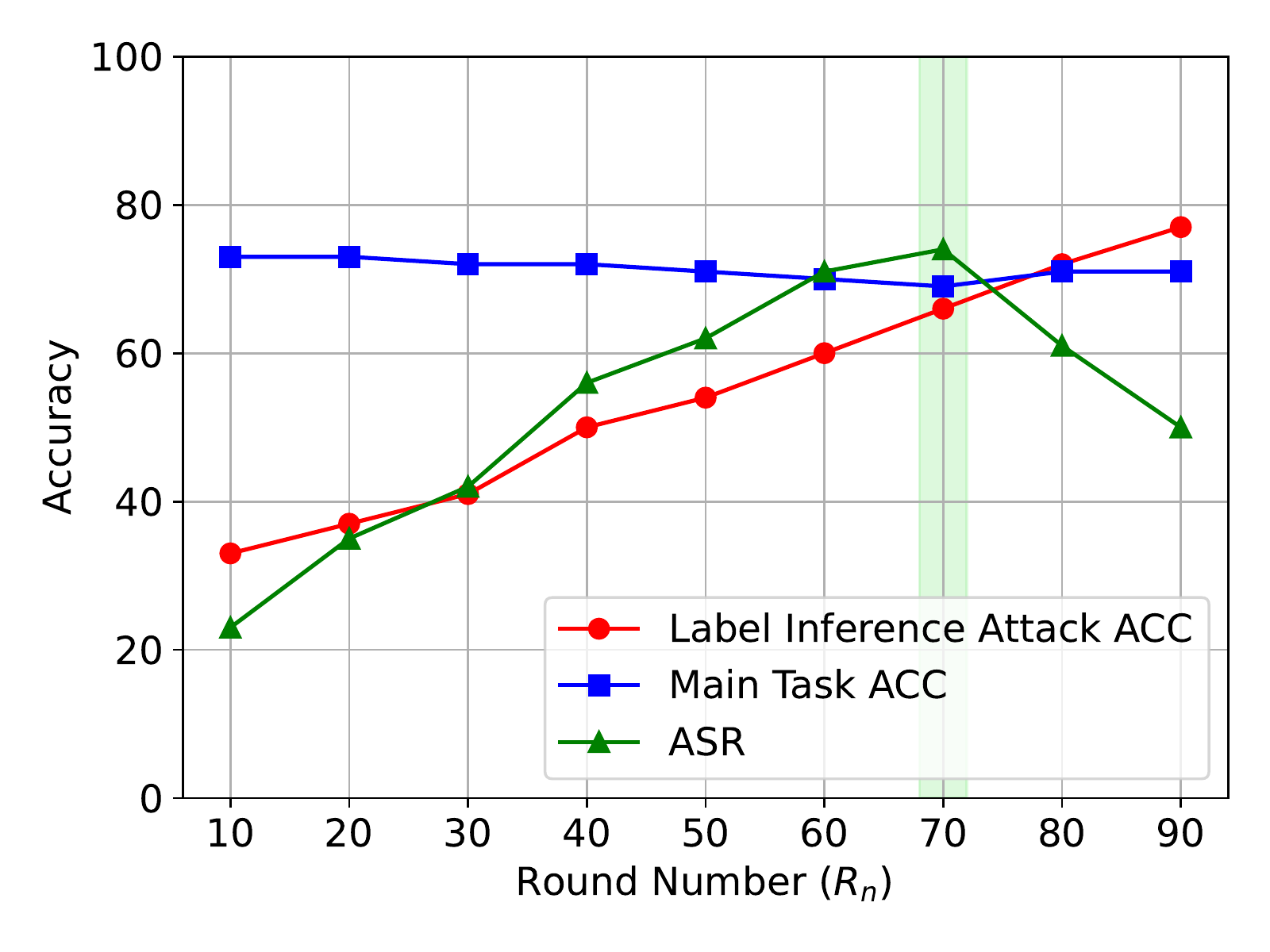}}\hspace*{-0.2cm}
\subfloat[Criteo]{\label{3metrics100rounds_2party_criteo}\includegraphics[width=0.255\textwidth]{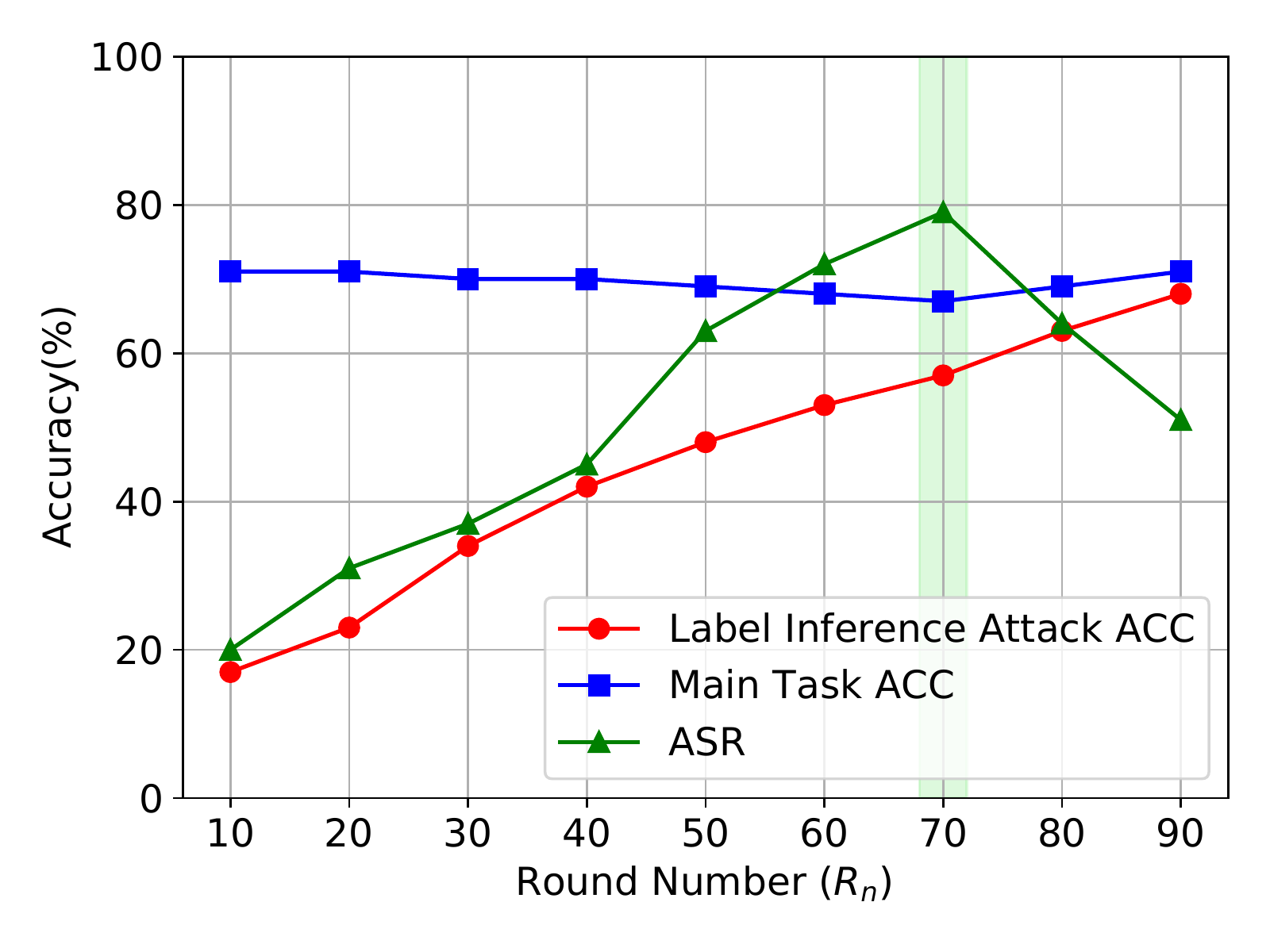}}\hspace*{-0.2cm}
\caption{\edits{Main Task Accuracy, label inference attack accuracy, and ASR for the two-party setting over different values of $R_n$, which denotes the point when the inference attack process ends and the backdoor attack begins. The poisoning budget is 10\%, sliding window is 5$\times$5 for image datasets, and source and target classes are randomly selected.}}
\label{3metrics100rounds_2party}
\end{figure*}

\begin{table}[t]
\centering
\resizebox{\columnwidth}{!}{
\begin{tabular}{ll|rr|rr|rr|rr}
\toprule
                         &               & \multicolumn{2}{c|}{\bf CIFAR-10}              & \multicolumn{2}{c}{\bf CIFAR-100}      & \multicolumn{2}{c}{\bf CINIC-10}    & \multicolumn{2}{c}{\bf Criteo}         \\ 
                         &               & \bf ASR & \bf MTA   & \bf ASR & \bf MTA  & \bf ASR  & \bf MTA & \bf ASR  & \bf MTA\\ \midrule
\multirow{8}{*}{\rotatebox[origin=c]{90}{\textbf{Round Number ($R_n$)}}} 
					   & 10    & 0.35      & 0.81      & 0.25      & 0.74  & 0.23     & 0.73    &0.20  & 0.71\\
                         & 20 	& 0.43      & 0.80         & 0.31     & 0.74  & 0.35      & 0.73 &0.31  & 0.71   \\
                         & 30          & 0.73         & 0.79     & 0.42     & 0.73    & 0.42      & 0.72 & 0.37  &  0.70 \\
                         & 40        & 0.75      & 0.78     & 0.52     & 0.73    & 0.56      & 0.72 &  0.45  & 0.70  \\
                         & 50          & 0.79      & 0.78     & 0.60     & 0.72    & 0.62      & 0.71 & 0.63  & 0.69  \\
                         & 60	 & \textbf{0.85}      & 0.77     & \textbf{0.78}     & 0.70    &  0.71    & 0.70 & 0.72  & 0.68  \\
                         & 70       & 0.80      & 0.78     & 0.76     & 0.71    & \textbf{0.74}       & 0.69 & \textbf{0.79}  & 0.67 \\
                         & 80          & 0.61  & 0.80 & 0.61 & 0.72 & 0.61      & 0.71 & 0.64  & 0.69 \\
                         & 90          & 0.37  & 0.81 & 0.47 & 0.72 & 0.50      & 0.71 & 0.51  & 0.71 \\
 \bottomrule
\end{tabular}
}
\caption{ASR and main task accuracy (MTA) with different numbers of rounds as per Fig.~\ref{3metrics100rounds_2party}.} 
\label{tab:backdoormaintaskinroundnumber}
\end{table}

\edits{
We experiment with a gradually decreasing poisoning budget of $p\%=50\%, 10\%, 5\%$, and $1\% $. 
We set the size of the sliding window between 3$\times$3 and 5$\times$5 pixel with all image datasets.
Fig.~\ref{backdooredimages} presents some examples of backdoored images. 
The location of the trigger in image datasets, is determined by the saliency map that was generated during training which differs for each data sample.
}

\descr{Choosing $\mathbf{R_n}$.} It is a challenging task to set the value $R_n$ (denoting the round when the label inference stops and the backdoor attack begins), aiming to determine the optimal threshold between label inference and backdoor attack phases. 
To address this issue, we conduct experiments with different rounds, treating each as a possible value for $R_n$, 
We also measure the main task accuracy after all rounds. 

\edits{
Fig.~\ref{3metrics100rounds_2party} reports the accuracy of the label inference attack, main task, and BadVFL's ASR, with the number of rounds rangings from 10 to 90.
Table~\ref{tab:backdoormaintaskinroundnumber} reports the ASR and main task accuracy.
The results suggest that selecting $R_n$ at approximately 60 yields the highest ASR for CIFAR-10 and CIFAR-100 and 70 for CINIC-10 and Criteo. 
They also show the impact of appropriate $R_n$ values over the attack performance; intuitively, it should be neither too large nor too small.
Larger $R_n$ values yield lower ASRs, as fewer training rounds are dedicated to tuning the bottom and top layer models to memorize the association between the trigger-embedded training samples and attacker-desired class labels.
Whereas smaller $R_n$ values do not provide accurate label estimation to the attacker, which significantly decreases ASR.
}

\begin{figure}[t]
\centering
\includegraphics[width=0.35\textwidth]{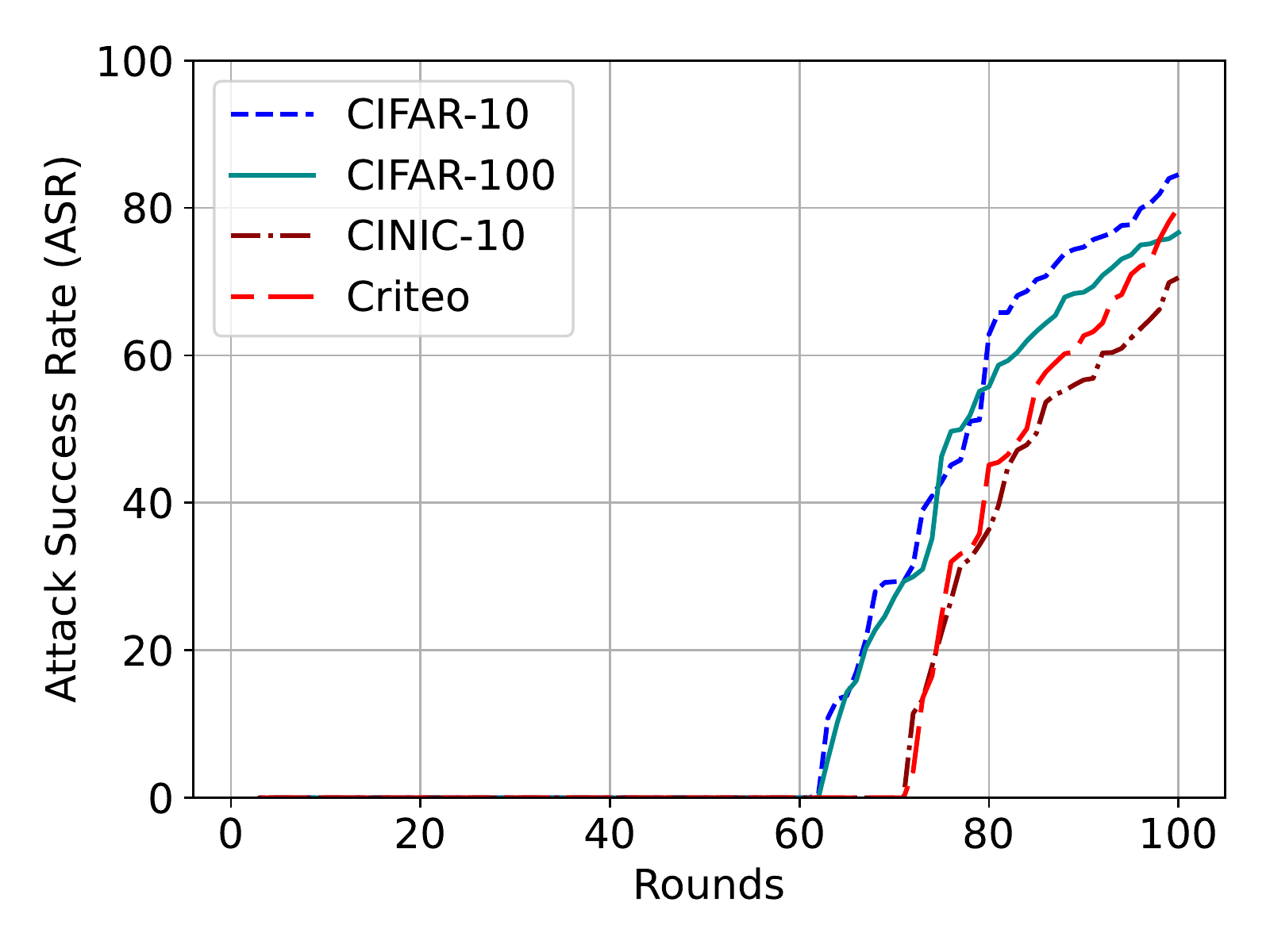}
\caption{Performance of the backdoor attack in different rounds, following the same setting as in Fig.~\ref{3metrics100rounds_2party}.}
\label{fig:ASRallthreedatasets}
\end{figure}

In our experiments, we empirically set $R_n$ so that the main learning accuracy of the VFL model reaches around 70\% of the testing accuracy obtained at the convergence of VFL training.
The idea is to let the VFL model close to convergence. 
This way, the model extraction and label estimate results are more stable and accurate than those obtained at the earlier stage of the training process.
Fig.~\ref{fig:ASRallthreedatasets} shows the ASR of the BadVFL attack continues to increase with more rounds dedicated to the backdoor poisoning stage of BadVFL.
Before the training process stops, using more rounds for poisoning can force the top model on the server to better fit the association between the backdoor trigger-embedded training samples and the desired class labels. 
\edits{
BadVFL is a clean-label backdoor attack process, with the key idea to increase the overlapping between the feature embeddings of the triggered source class data samples and those of the target class samples, which causes a drop in the utility of the main learning task.
}

\descr{Source and Target Class Selection.}
We experiment with the \textit{Optimal Selection} approach to select the source and target classes, having the adversary compute the pairwise similarity relation between classes using feature embeddings. 
The source and target classes are chosen once the label estimates are obtained, i.e., after 60 and 70 rounds, respectively, for CIFAR-10/CIFAR-100, and CINIC-10.

\begin{table}[t]
\centering
\begin{tabular}{l@{}r|rr|rr}
\toprule
\textbf{Dataset} & \multirow{1}{*}{\textbf{Poisoning}} & \multicolumn{2}{c|}{\textbf{\em Random Selection}}     & \multicolumn{2}{c}{\textbf{\em Optimal Selection}}    \\
      &      \bf{Budget}                   & \bf{ASR} & \textbf{MTA} & \bf{ASR} & \textbf{MTA} \\ \midrule

CIFAR-10   & \multirow{3}{*}{10\%}       & 0.85   &         0.77       & 0.89  &   0.76             \\ 
CIFAR-100   &      & 0.78    &       0.70         & 0.81   &    0.67            \\ 
CINIC-10    &      & 0.74   &      0.69          &  0.77  &    0.67            \\
\midrule
CIFAR-10 & \multirow{3}{*}{5\%}         & 0.72   &        0.80       & 0.84  &  0.81             \\ 
CIFAR-100&     & 0.64    &      0.73         & 0.78   &   0.71            \\ 
CINIC-10 &   & 0.61   &     0.72          &  0.72  &   0.71            \\
\bottomrule
\end{tabular}
\caption{Random vs.~Optimal Selection of source and target classes using two poisoning budgets.} 
\label{tab:randomvsoptimalselection}
\vspace{0.2cm}
\end{table}

Table~\ref{tab:randomvsoptimalselection} presents the results of the experiments. 
The poisoning budget and the window size are the same for both selections.
The (source class, target class) pairs selected at random for the CIFAR-10, CIFAR-100, and CINIC-10 datasets are (airplane, automobile), (bus, train), and (ship, truck), respectively.
However, the (source class, target class) pairs selected for optimal performance for the CIFAR-10, CIFAR-100, and CINIC-10 datasets are (dog, cat), (bee, beetle), and (dog, cat), respectively.
The optimal selection approach chooses the source and target class that stay close to the feature embedding space.
This facilitates the adversary: the closer the inter-class distance is, the easier it is for her to make the perturbed instances of the target class closers to the triggered instances of the source class.
\edits{
Note that the optimal selection is not applicable in Criteo dataset as there are only two classes.
}

From Table~\ref{tab:randomvsoptimalselection}, we observe a significant improvement in BadVFL's ASR with Optimal Selection. 
We also find that the main task accuracy (MTA) is slightly lower than that of the Random Selection setting. 
The clean-label attack strategy in BadVFL increases the overlapping between the source and target class in the feature space. 
On the one hand, Optimal Selection causes more inter-class overlapping than Random Selection, which strengthens the attack.
On the other hand, it unavoidably causes confusion in the decision boundary of the target classifier, which results in a drop in MTA. 
Nevertheless, using Optimal Selection increases ASR more than the loss in MTA over the three datasets. 
This confirms the benefits of using Optimal Selection in boosting the attack performances of BadVFL.

\begin{table}[t]
\centering
\setlength{\tabcolsep}{4pt}
\begin{tabular}{r|rr|rr|rr|rr}
\toprule
       \textbf{Poisoning}                               & \multicolumn{2}{c|}{\bf CIFAR-10}              & \multicolumn{2}{c|}{\bf CIFAR-100}      & \multicolumn{2}{c}{\bf CINIC-10}       & \multicolumn{2}{c}{\bf Criteo}      \\ 
\textbf{Budget}                                       & \bf ASR & \bf MTA   & \bf ASR & \bf MTA  & \bf ASR  & \bf MTA & \bf ASR  & \bf MTA\\ \midrule

					    50\%    & 0.93      & 0.70      & 0.84      & 0.60  & 0.82     & 0.62  & 0.88     & 0.63 \\
10\% 	& 0.85     & 0.77         & 0.78     & 0.70  & 0.74      & 0.69   & 0.79     & 0.67 \\
5\%          & 0.72         & 0.80     & 0.64     & 0.71    & 0.61      & 0.71 & 0.70     & 0.73 \\
                          1\%        & 0.61      & 0.81     & 0.47     & 0.73    & 0.55      & 0.73 & 0.58     & 0.74 \\

 \bottomrule
\end{tabular}
\caption{ASR and main task accuracy (MTA) with different poisoning budgets in BadVFL.} 
\label{tab:poisoningbudget}
\end{table}

Moreover, Table~\ref{tab:randomvsoptimalselection} shows that by using the Optimal Selection approach, we can reduce the poisoning budget and achieve a better main task accuracy with a similar ASR. 
For example, in the CIFAR-100 setting with a 5\% poisoning budget, using Optimal Selection produces an ASR of 78\% compared to the 10\% poisoning budget in the random selection approach.

\descr{Poisoning Budget.} 
We also vary the poisoning budget and present the corresponding results in Table~\ref{tab:poisoningbudget}.
When increasing the poisoning budget, the effectiveness of the backdoor attack (i.e., the ASR) increases accordingly. 
This is consistent with the design principle of the BadVFL-based backdoor attack. 
More perturbed training instances injected into the training data set of the target class can cause more significant overlapping between the source and target class in the feature embedding space. 
The increasingly larger overlapping help boost ASR. 
However, main task accuracy decreases at the same time. 
The reason is that increasing inter-class overlapping between the source and target class makes the two classes more difficult to differentiate, which leads to more misclassification. 
Also note that, in a two-party setting, the adversary holds half of the features and has an equal contribution as the benign participant; this explains why the poisoning budget has a significant impact on the accuracy of the main learning task.

\descr{How to embed the trigger.}
We first experiment with a varying size of the injected trigger in image datasets. 
We evaluate the attack performance using two different sliding window sizes; see Table~\ref{tab:slidingwindow}. 
We observe that increasing the size of the trigger will slightly decrease the accuracy of the main task. 
As the adversary sets up an association between the trigger and the label, a larger size of the trigger can bring a stronger signal to learn the association and eventually increases the ASR. 

We also focus on the effectiveness of using a saliency map to guide the trigger embedding step. 
The loss of the target classifier is more prone to the perturbation added to the image areas of high salient values. 
In theory, embedding trigger signals to the highly salient areas causes a large change in the classification boundary and yields high ASR. 

We experimentally confirm the benefits of using the saliency map, %
and report the resulting ASR/MTA on CIFAR-10 in Table~\ref{tab:saliencyvsrandom} when the trigger is inserted randomly into the selected training images without using a saliency map.
In this experiment, the attacker does not rely on the saliency map to insert the backdoor. 
Rather, she randomly selects an area to insert the backdoor.
To ensure a fair comparison, we let the adversary execute the attack five times using different random areas and then compute ASR and MTA.

When comparing the results in Table~\ref{tab:saliencyvsrandom} to those using the saliency map in Table~\ref{tab:slidingwindow}, it is clear that the attack worsens when the trigger is added randomly into the data sample. 
By contrast, using the saliency map boosts ASR from 0.62 to 0.85, %
with a marginal impact on the accuracy of the main task (MTA dropping slightly from 0.78 to 0.77). 

In machine learning research, saliency maps are also known as axiomatic attribution measurements of classifiers. 
According to previous work~\cite{Chen2022nips,Yang2022acl}, although the salient areas are sensitive to noise perturbation, 
these areas of high saliency values do not necessarily contribute the most to the overall classification performances 
Thus, perturbing these areas does not necessarily reduce overall accuracy.

\begin{table}[t]
\centering
\setlength{\tabcolsep}{5pt}
\begin{tabular}{r|rr|rr|rr}
\toprule
                  {\bf Sliding}                    & \multicolumn{2}{c|}{\bf CIFAR-10}              & \multicolumn{2}{c|}{\bf CIFAR-100}      & \multicolumn{2}{c}{\bf CINIC-10}            \\ 
{\bf    Window Size}    & \bf ASR & \bf MTA   & \bf ASR & \bf MTA  & \bf ASR  & \bf MTA\\ \midrule
3$\times$3    & 0.79      & 0.80      & 0.72      & 0.73  & 0.70     & 0.71   \\
5$\times$5		& 0.85     & 0.77         & 0.78     & 0.70  & 0.74      & 0.69    \\
 \bottomrule
\end{tabular}
\caption{ASR and main task accuracy (MTA) in image datasets with different sliding window sizes (other factors are same as in Fig.~\ref{3metrics100rounds_2party}).} 
\label{tab:slidingwindow}
\end{table}

\begin{table}[t]
\centering
\begin{tabular}{l|cc|cc}
\toprule
{\bf Dataset} & \multicolumn{2}{c|}{\textbf{Saliency Map}}     & \multicolumn{2}{c}{\textbf{Random Insertion}}    \\
                  & \multicolumn{1}{c}{\textbf{ASR}} & \textbf{MTA} & \multicolumn{1}{c}{\textbf{ASR}} & \textbf{MTA} \\ \midrule
CIFAR-10          & 0.85  &         0.77       & 0.62  &   0.78             \\ 
\bottomrule
\end{tabular}
\caption{ASR and main task accuracy (MTA) with an attacker employing a saliency map vs.~randomly inserting the backdoor.} 
\label{tab:saliencyvsrandom}
\vspace{0.2cm}
\end{table}

\descr{Overlapping Auxiliary and Training Data Space.}
\edits{
In the previous experiments, we assumed that the auxiliary dataset shares the same label space as the true training data.
Next, we study the impact of reducing the overlapping percentage of the label space between the auxiliary and training dataset on the performance of our backdoor attack.
We assume the former only contains the labeled data points belonging to $K$ classes out of total $N$ and unlabeled data points not belonging to these $K$ classes.
We train the label inference module classifier so that it can recognize samples from all of the $K$ classes.
However, any samples beyond the $K$ classes will be considered ``unrecognized class'' by the classifier.
We then run the new classifier on the training data to infer the labels and initiate the backdoor attack.
}

\begin{table}[t]
\centering
\begin{tabular}{lc|rrr}
\toprule
\textbf{Dataset} & \multirow{1}{*}{\textbf{Overlapping (\%)}} &\textbf{ASR}   & \textbf{MTA} & \textbf{LIA} \\ \midrule

CIFAR-10   & \multirow{2}{*}{70\%}       &   0.85       & 0.76  &   0.55             \\ 
CIFAR-100   &      & 0.77   & 0.70   &    0.41            \\ 
\midrule
CIFAR-10 & \multirow{2}{*}{40\%}         &0.84       & 0.76  &  0.34             \\ 
CIFAR-100&     & 0.75    & 0.69   &   0.32            \\ 
\bottomrule
\end{tabular}
\caption{\edits{ASR, main task accuracy (MTA), and label inference accuracy (LIA) with 70\% and 40\% overlap between the attacker's auxiliary data and training space.}}
\label{tab:overlapauxtraincifar}
\vspace{0.2cm}
\end{table}

\edits{
To this end, we perform experiments with $K=7$ and $K=4$ on the CIFAR-10 and CIFAR-100 datasets, with a total of $10$ and $100$ classes, respectively, in a two-party setting.
The results are presented in Table~\ref{tab:overlapauxtraincifar}.
We observe that an overlapping rate of less than 100\% does not impact the attack performance.
The main reason is that, in BadVFL, the adversary only needs to identify samples from the source and target classes, respectively.
If the samples from the source and target classes appear in the overlapped classes, the adversary can perform the attack.
This indicates that even if the adversary collects a significant amount of auxiliary data, the effectiveness of the backdoor attack can persist if she only assigns labels to certain classes.
}

\subsection{Multi-Party Experiments}
Finally, we consider more than two participants, ranging from 4 to 10, to evaluate the performance of BadVFL in a multi-party setting.
We first consider the Single Attacker Mode, where only one of the participants is controlled by the adversary and executes the backdoor attack. 
Then, we do so with the Multi-Attacker Mode: multiple participants are compromised by the adversary and collude to launch the attack. 
In both cases, the adversary executes the attack in two steps: label inference and backdoor trigger injection.

\subsubsection{Single Attacker mode}
We expect that with more participants in VFL, the impact of the poisoning attacks becomes weaker. %
This implies that the label inference and the backdoor trigger injection steps become less significant with only one of the $K$ participants. 
To guarantee that the backdoor poisoning effort is strong enough to deliver a successful attack, we assume the adversary uses a sliding window of size $5\times5$ to inject the trigger in the image datasets. 
We set the poisoning budget ($p\%$) to $20\%$ and use Optimal Selection to choose the source and target classes for the BadVFL attack.

\begin{figure*}[t]
\setlength{\tempwidth}{.25\linewidth}
\centering
\subfloat[CIFAR-10, 4/6 participants]{\label{CIFAR-10-4-6}\includegraphics[width=\tempwidth]{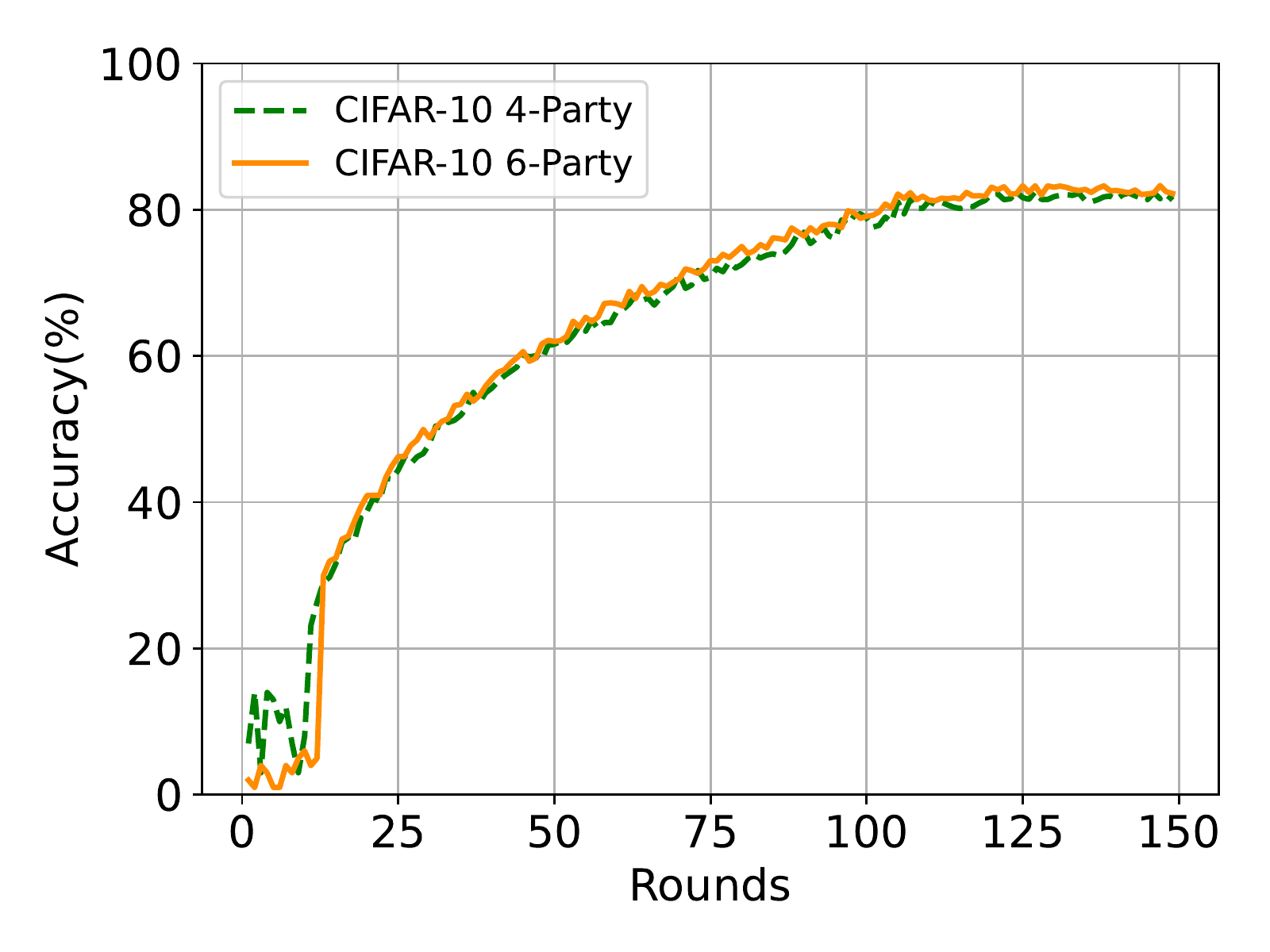}}
\subfloat[CIFAR-100, 4/6 participants]{\label{CIFAR-100-4-6}\includegraphics[width=\tempwidth]{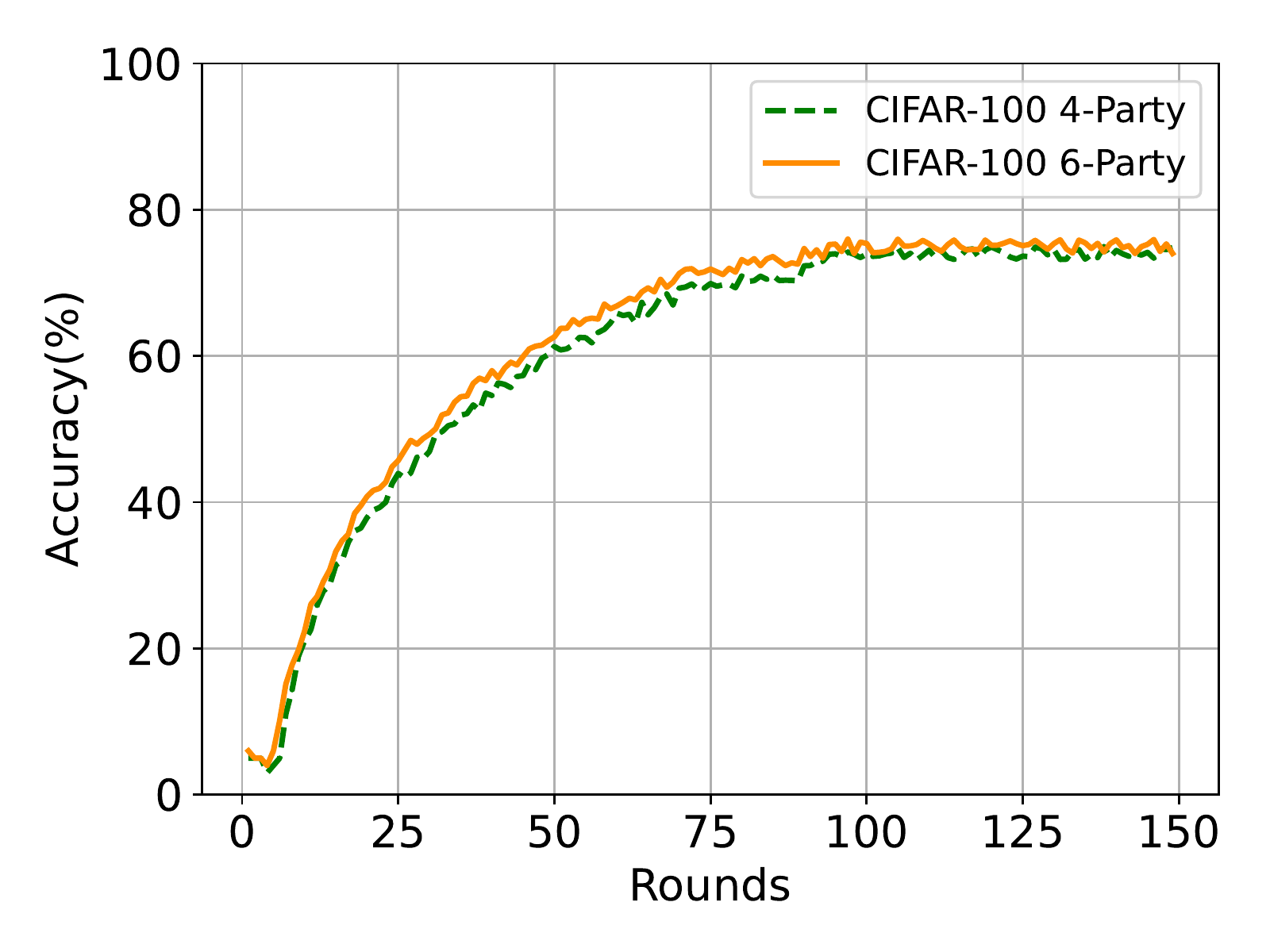}}
\subfloat[CINIC-10, 4/6 participants]{\label{CINIC-10-4-6}\includegraphics[width=\tempwidth]{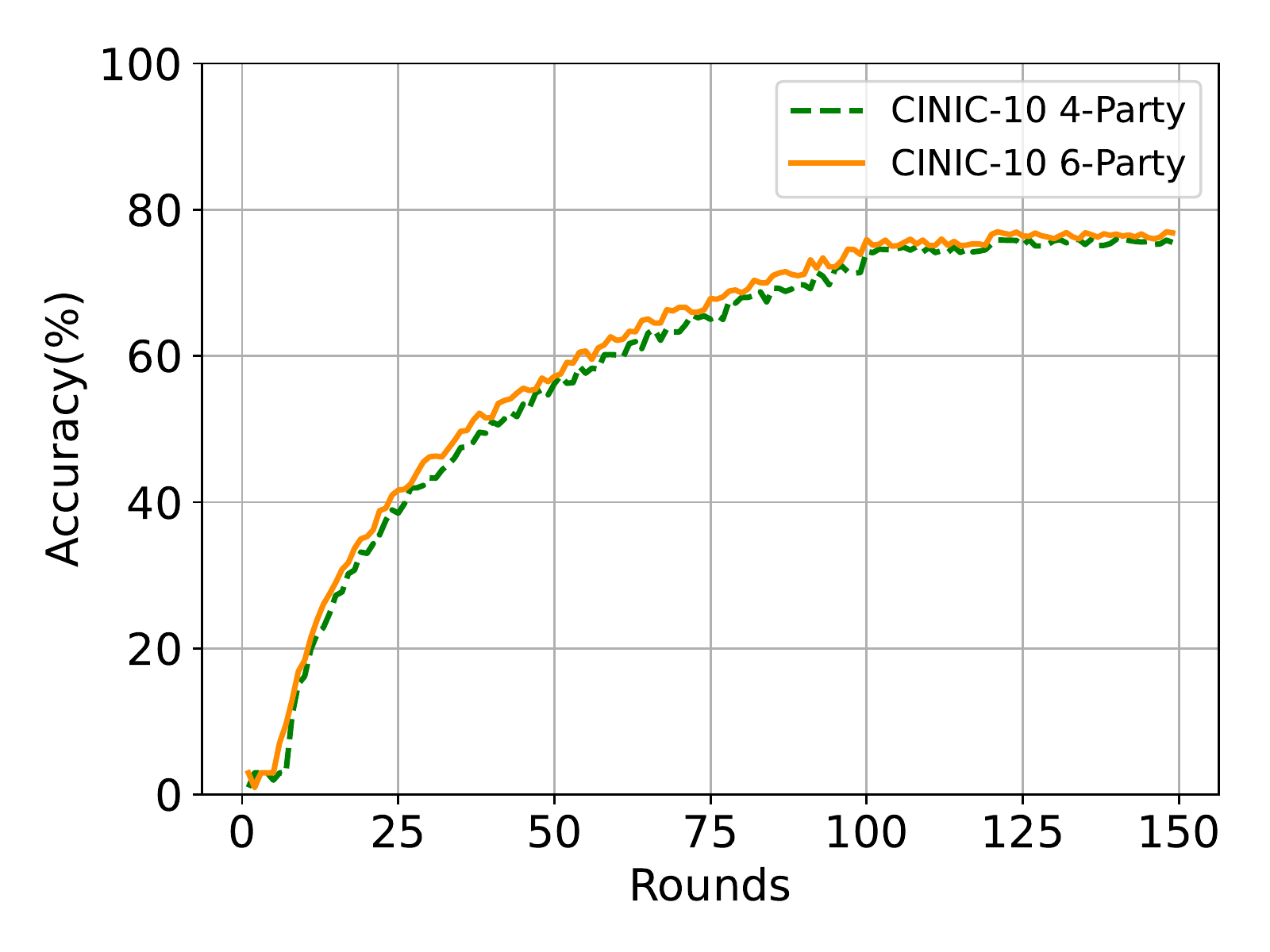}}
\subfloat[Criteo, 4/6 participants]{\label{Criteo-10-4-6}\includegraphics[width=\tempwidth]{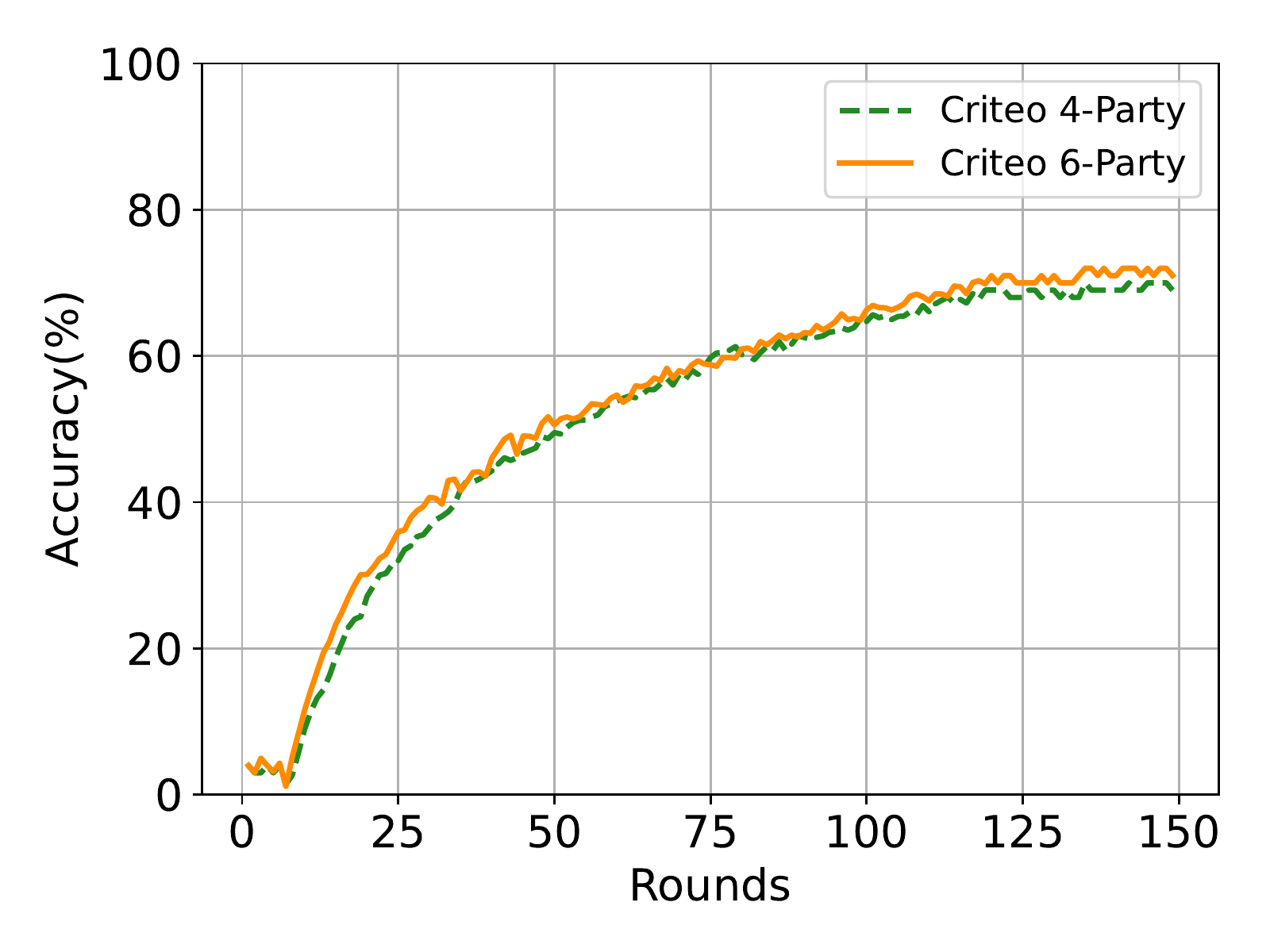}}\\
[-2ex]
\subfloat[CIFAR-10, 8/10 participants]{\label{CIFAR-10-8-10}\includegraphics[width=\tempwidth]{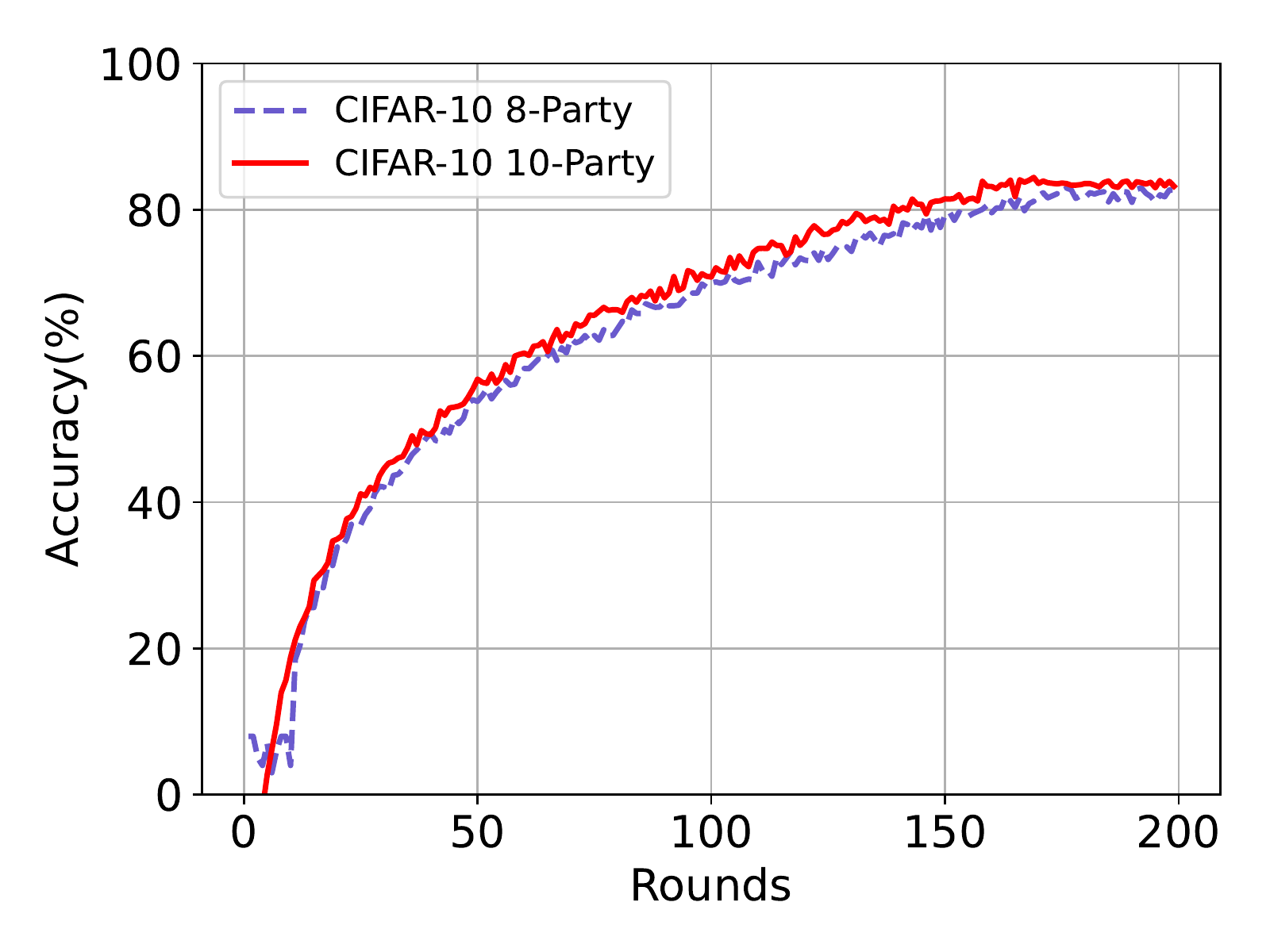}}
\subfloat[CIFAR-100, 8/10 participants]{\label{CIFAR-100-8-10}\includegraphics[width=\tempwidth]{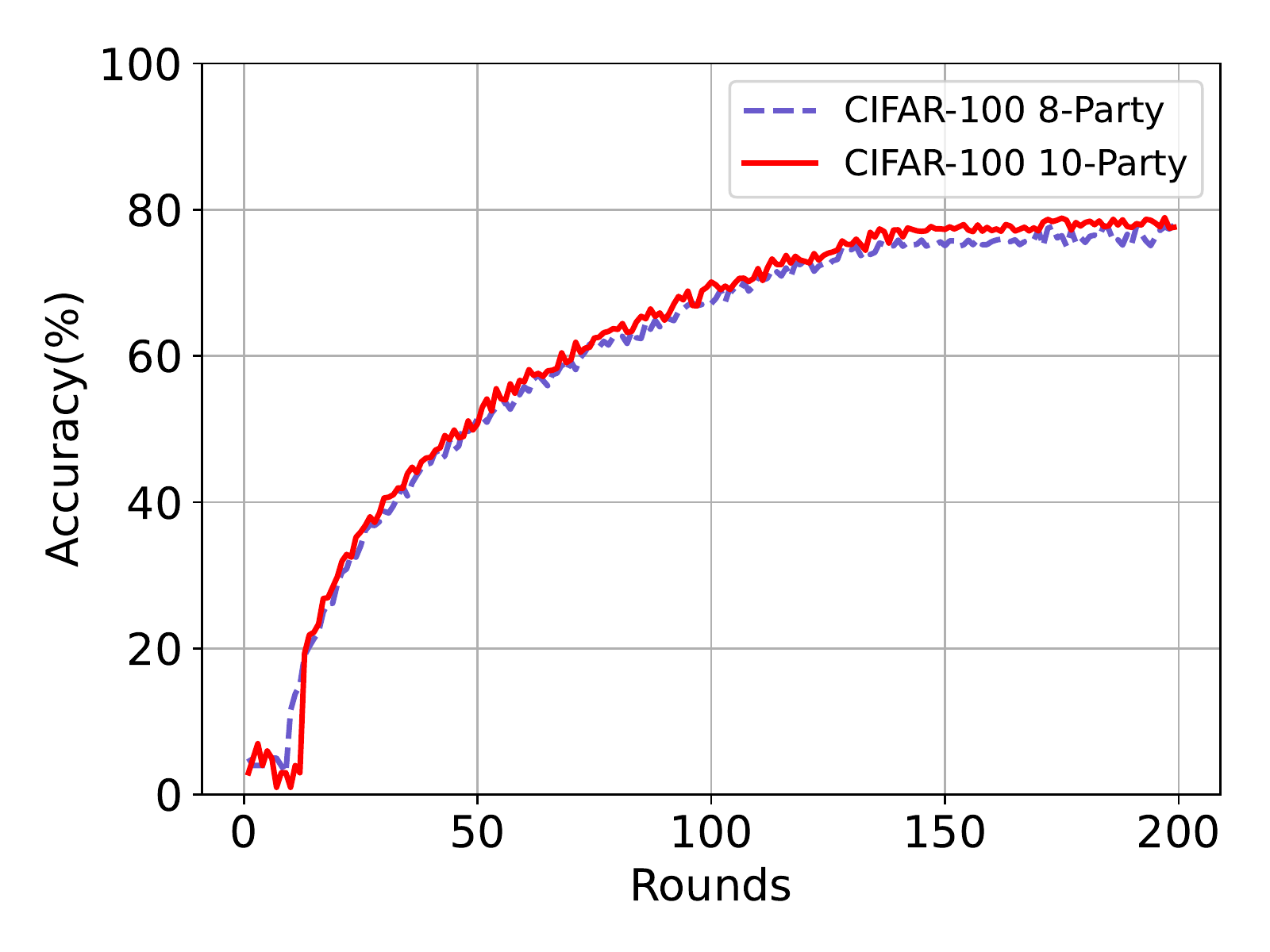}}
\subfloat[CINIC-10, 8/10 participants]{\label{CINIC-10-8-10}\includegraphics[width=\tempwidth]{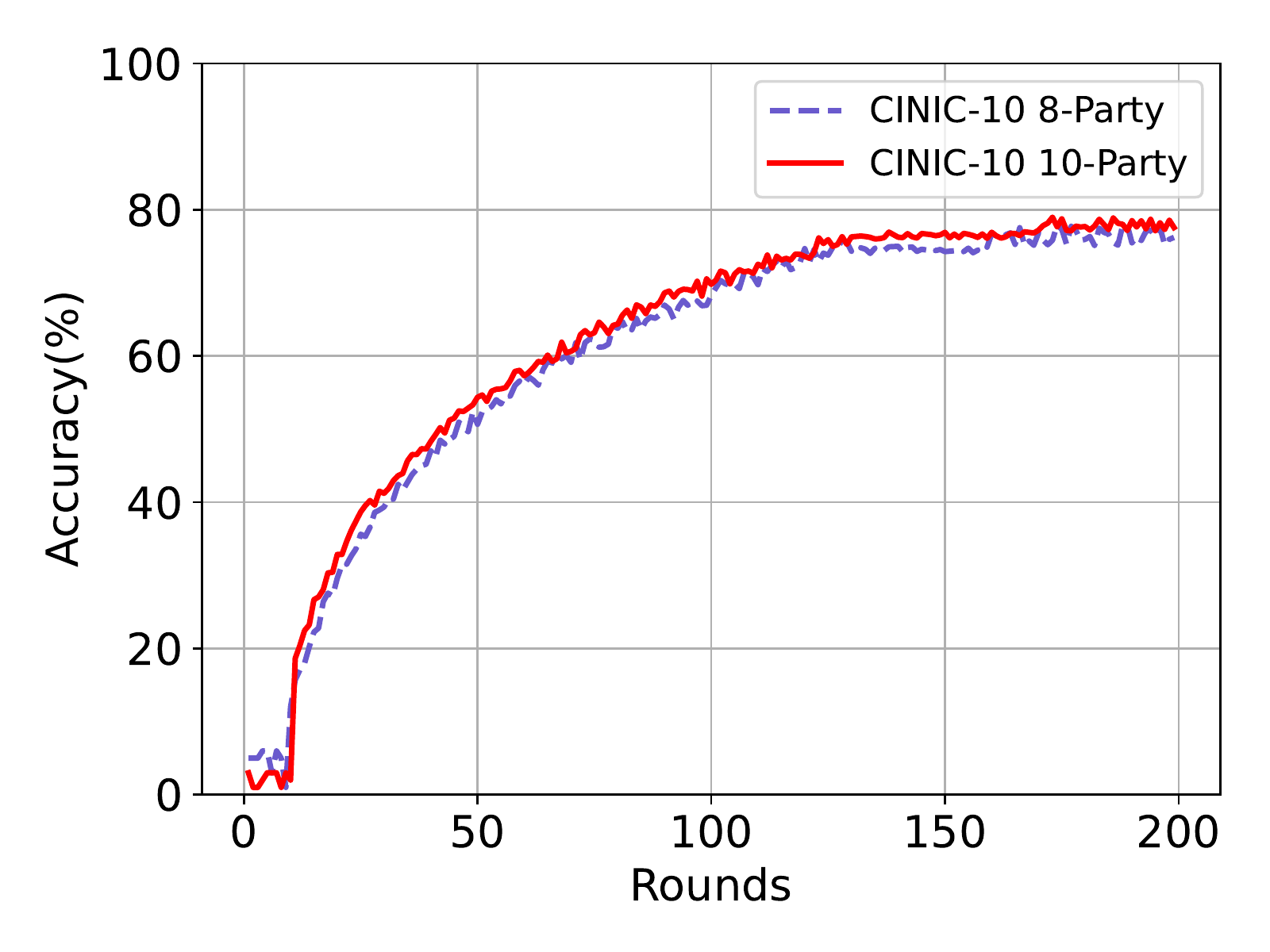}}
\subfloat[Criteo, 8/10 participants]{\label{Criteo-10-8-10}\includegraphics[width=\tempwidth]{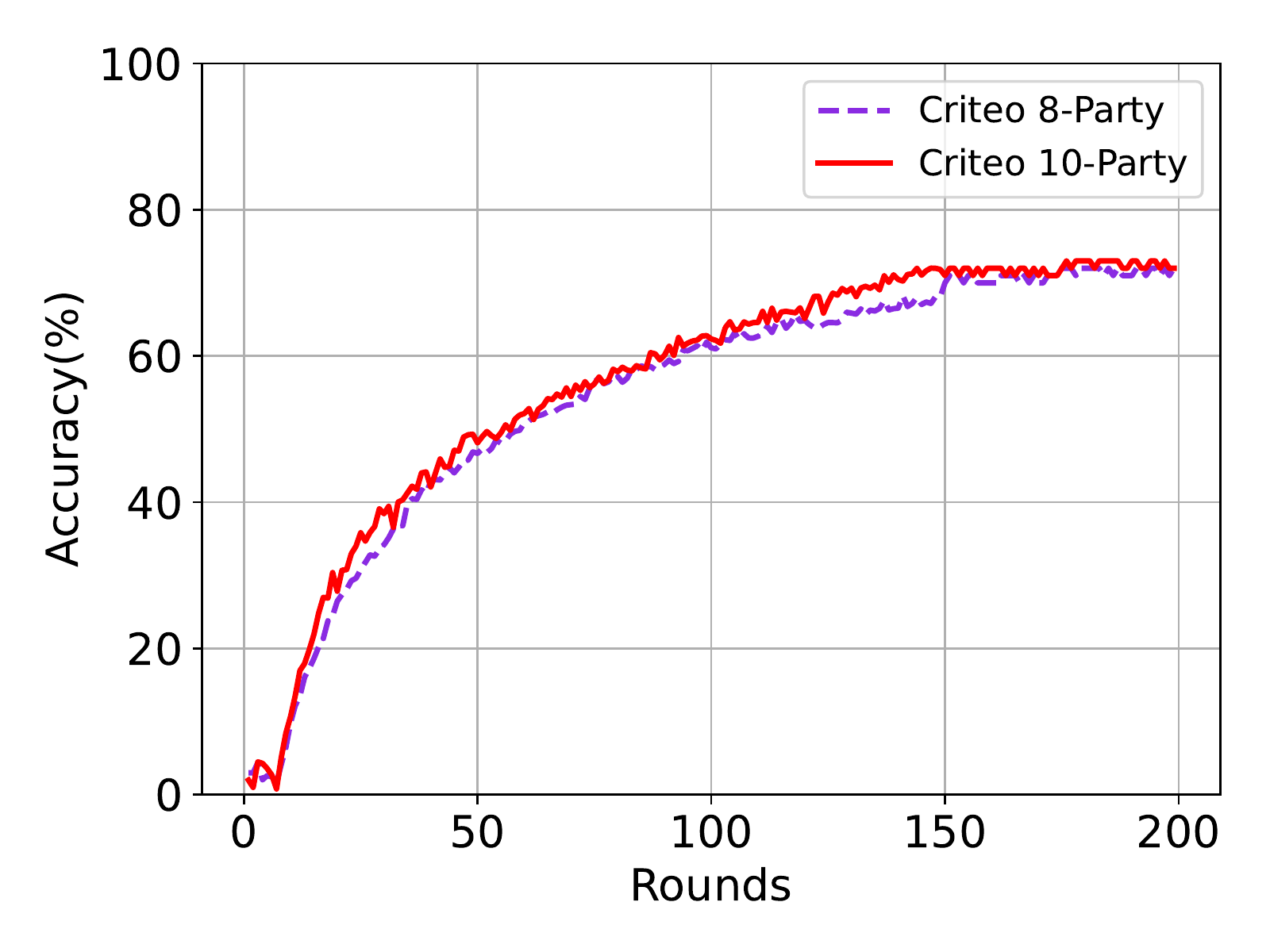}}
\caption{Main task accuracy (MTA) under no attack, in 150 rounds for 4-party and 6-party and in 200 rounds for 8-party and 10-party.}
\label{fig:maintaskmultiparty}
\end{figure*}

\begin{figure*}[t]
\setlength{\tempwidth}{.25\linewidth}
\centering
\subfloat[CIFAR-10, 4 participants]{\label{CIFAR-10-4party}\includegraphics[width=\tempwidth]{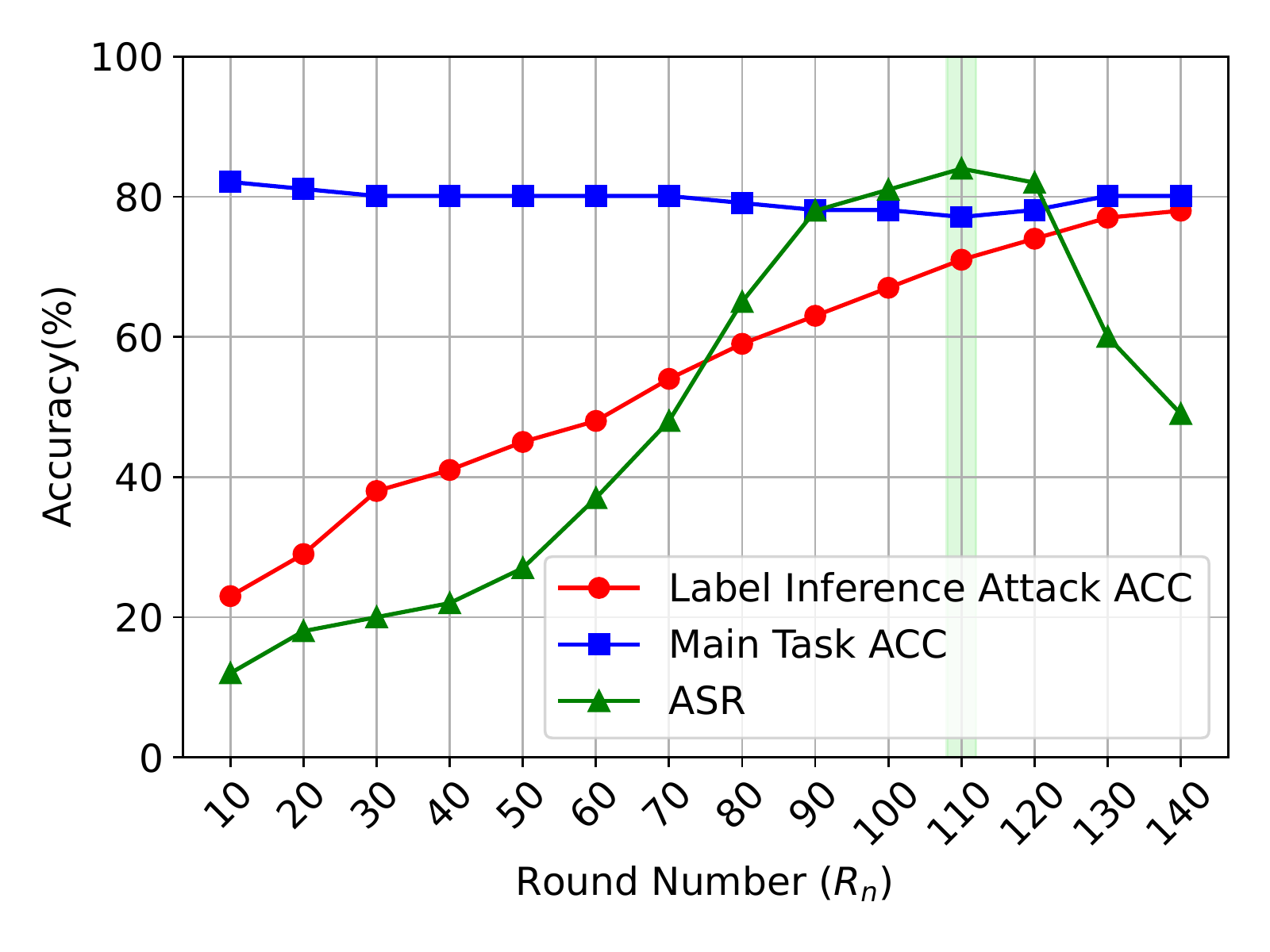}}
\subfloat[CIFAR-10, 6 participants]{\label{CIFAR-10-6party}\includegraphics[width=\tempwidth]{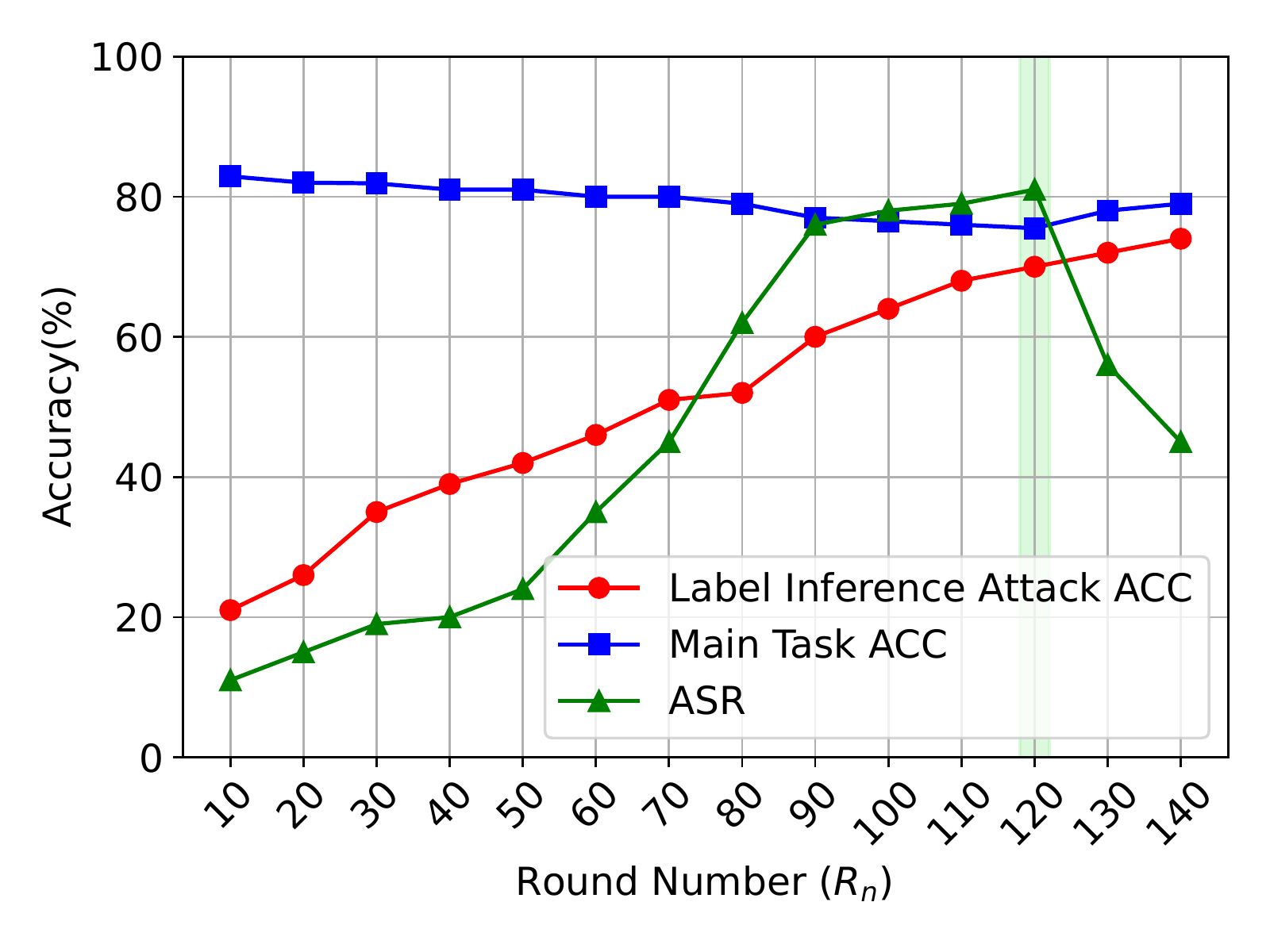}}
\subfloat[CIFAR-10, 8 participants]{\label{CIFAR-10-8party}\includegraphics[width=\tempwidth]{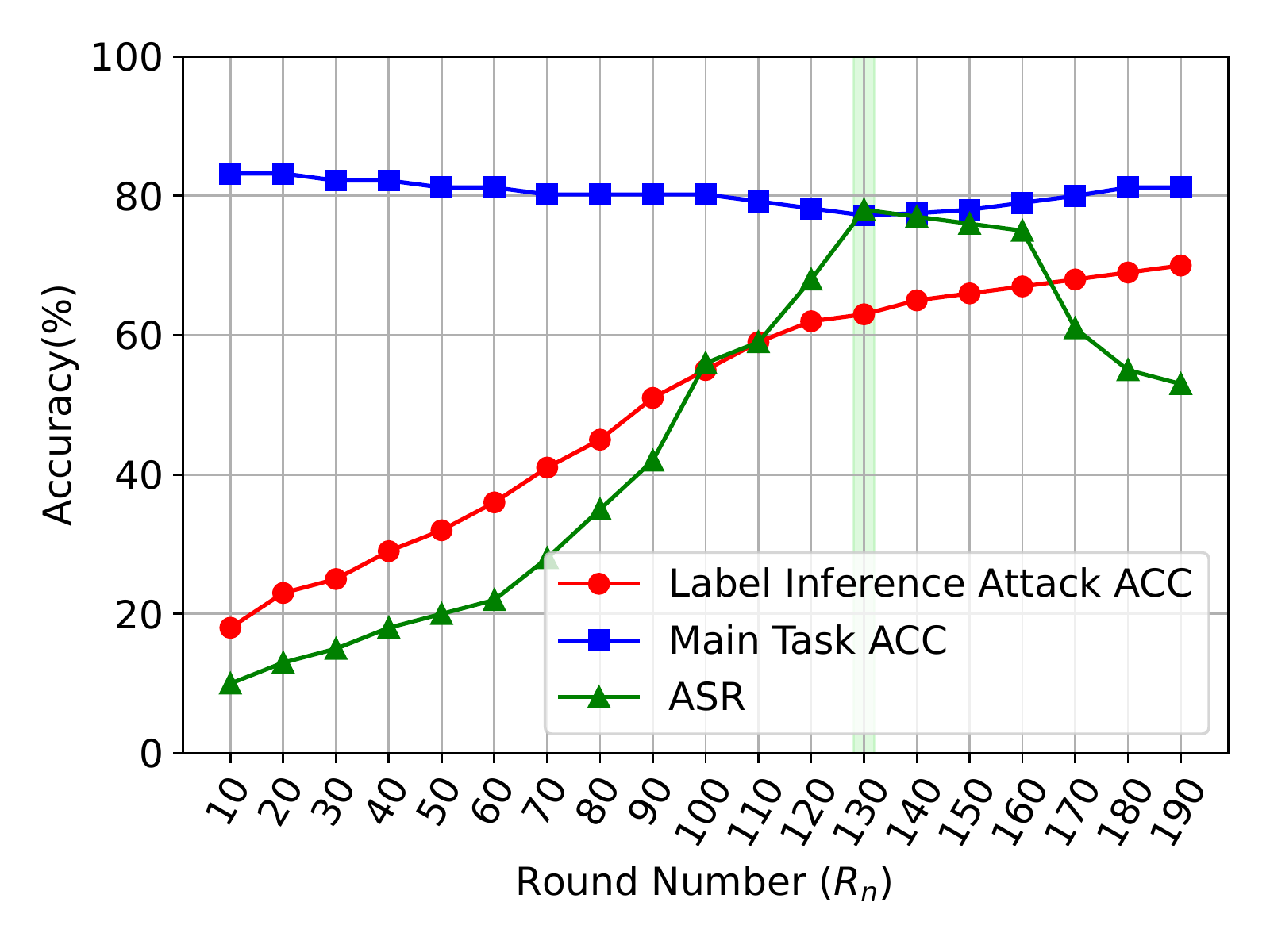}}
\subfloat[CIFAR-10, 10 participants]{\label{CIFAR-10-10party}\includegraphics[width=\tempwidth]{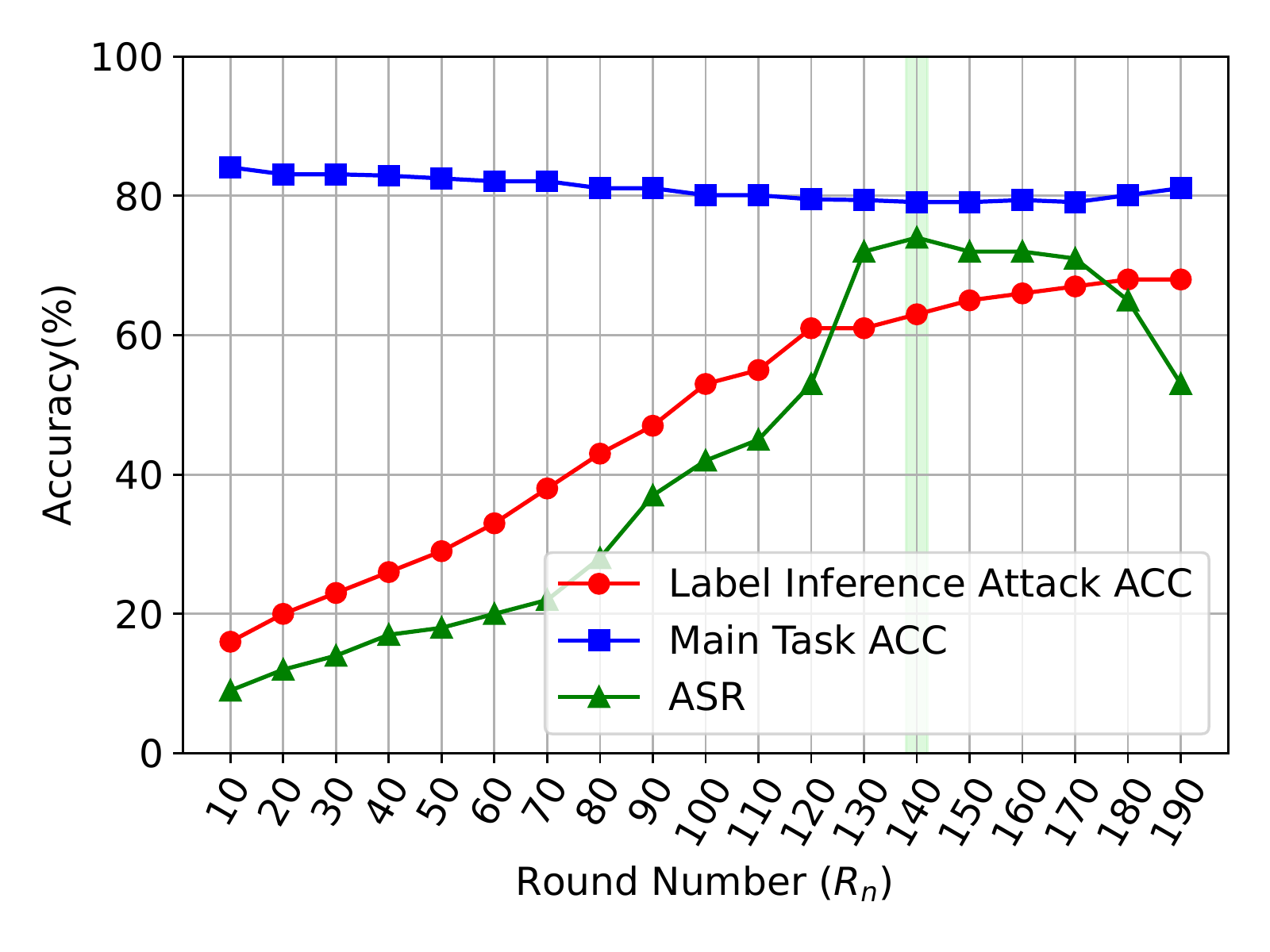}}\\
[-2ex]
\subfloat[CIFAR-100, 4 participants]{\label{CIFAR-100-4party}\includegraphics[width=\tempwidth]{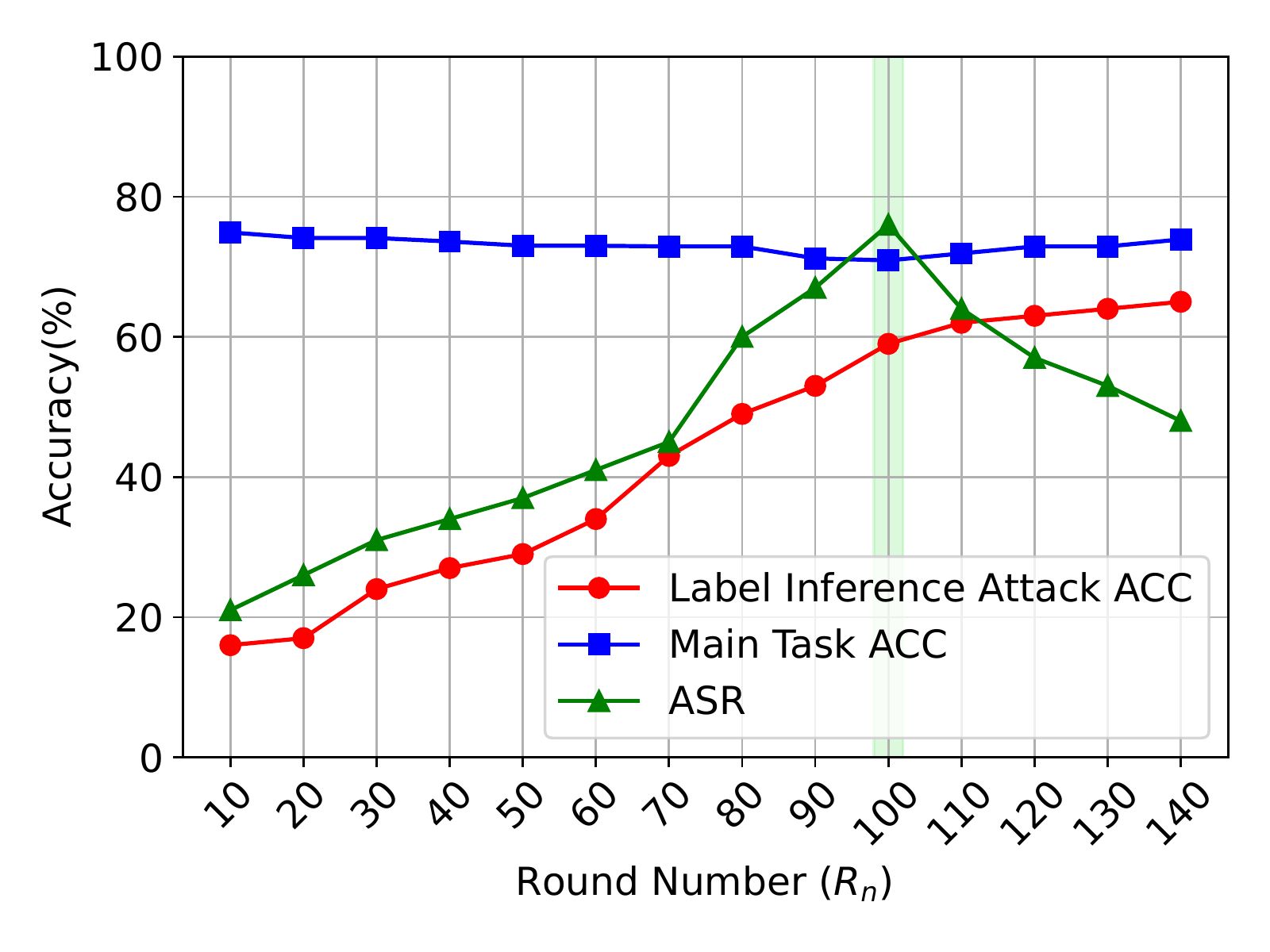}}
\subfloat[CIFAR-100, 6 participants]{\label{CIFAR-100-6party}\includegraphics[width=\tempwidth]{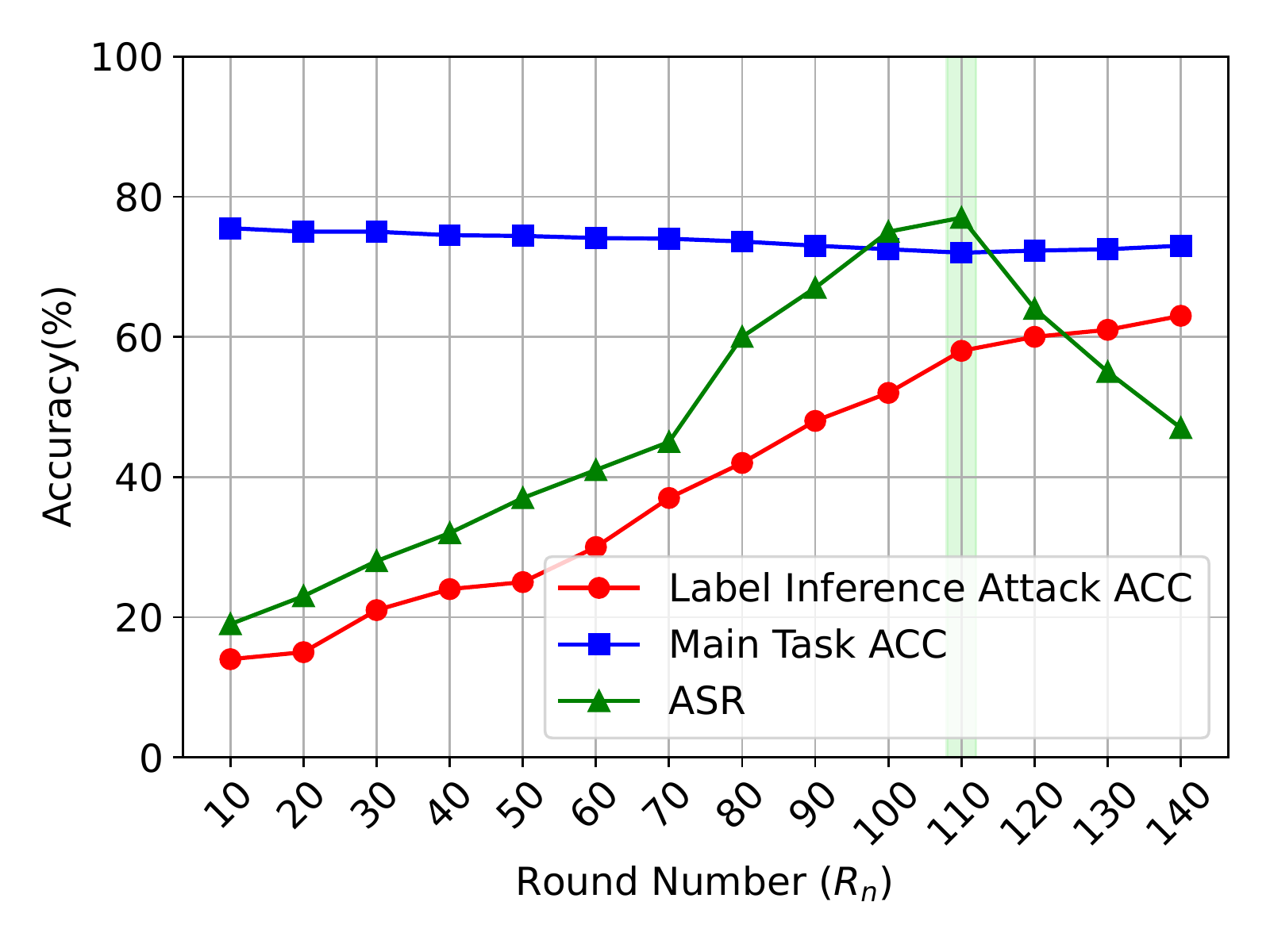}}
\subfloat[CIFAR-100, 8 participants]{\label{CIFAR-100-8party}\includegraphics[width=\tempwidth]{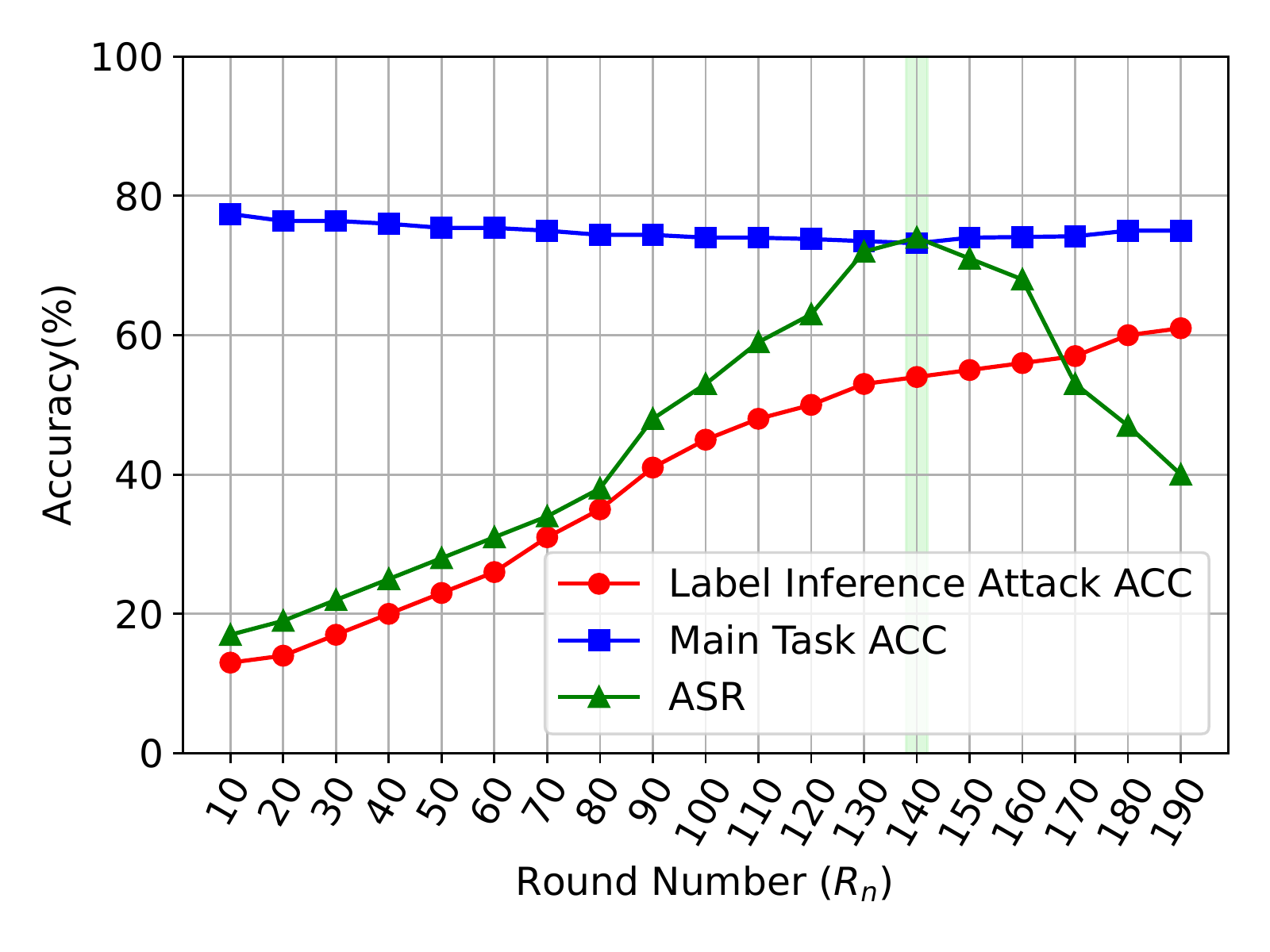}}
\subfloat[CIFAR-100 with 10 participants]{\label{CIFAR-100-10party}\includegraphics[width=\tempwidth]{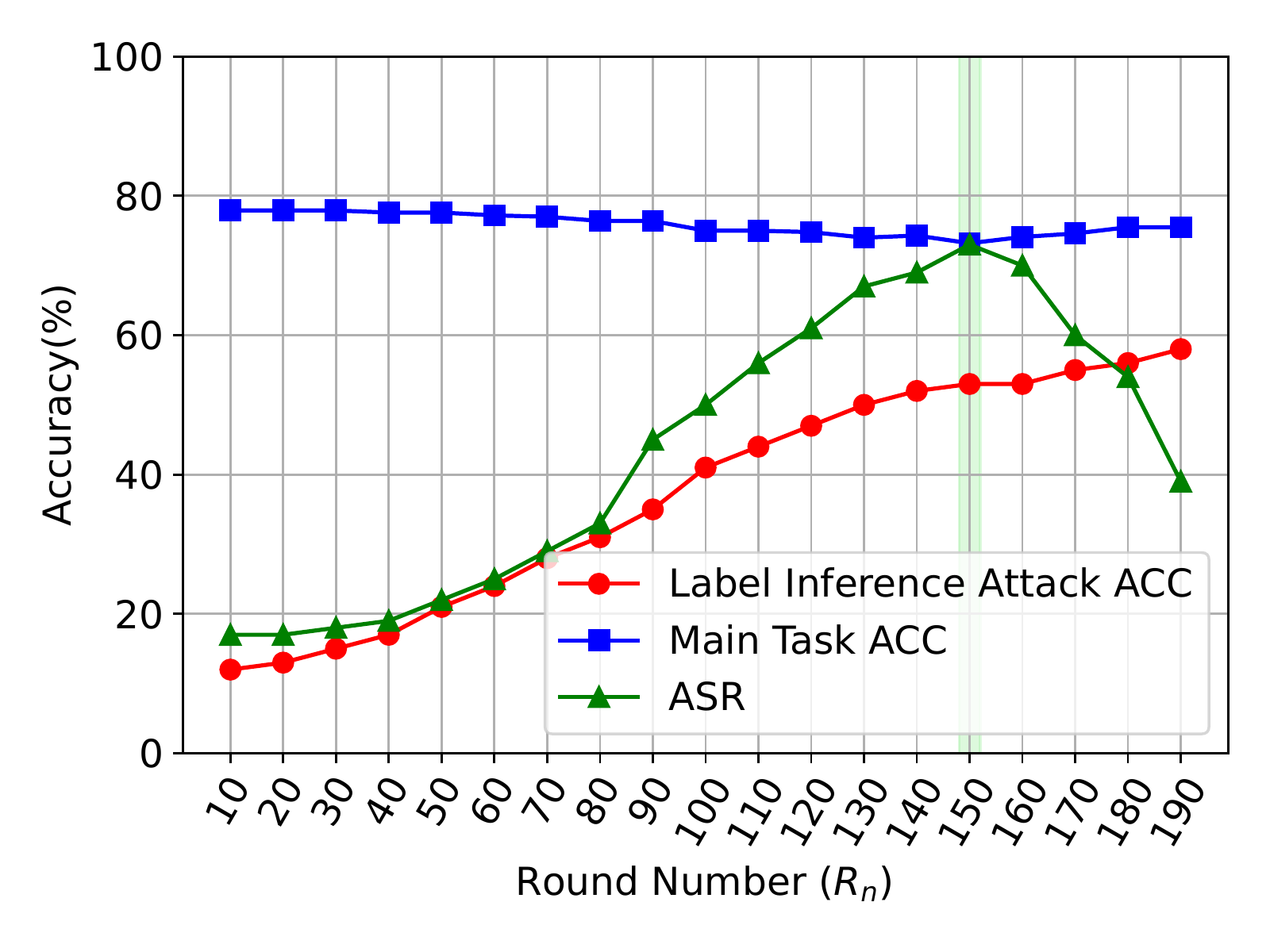}}\\
[-2ex]
\subfloat[CINIC-10, 4 participants]{\label{CINIC-10-4party}\includegraphics[width=\tempwidth]{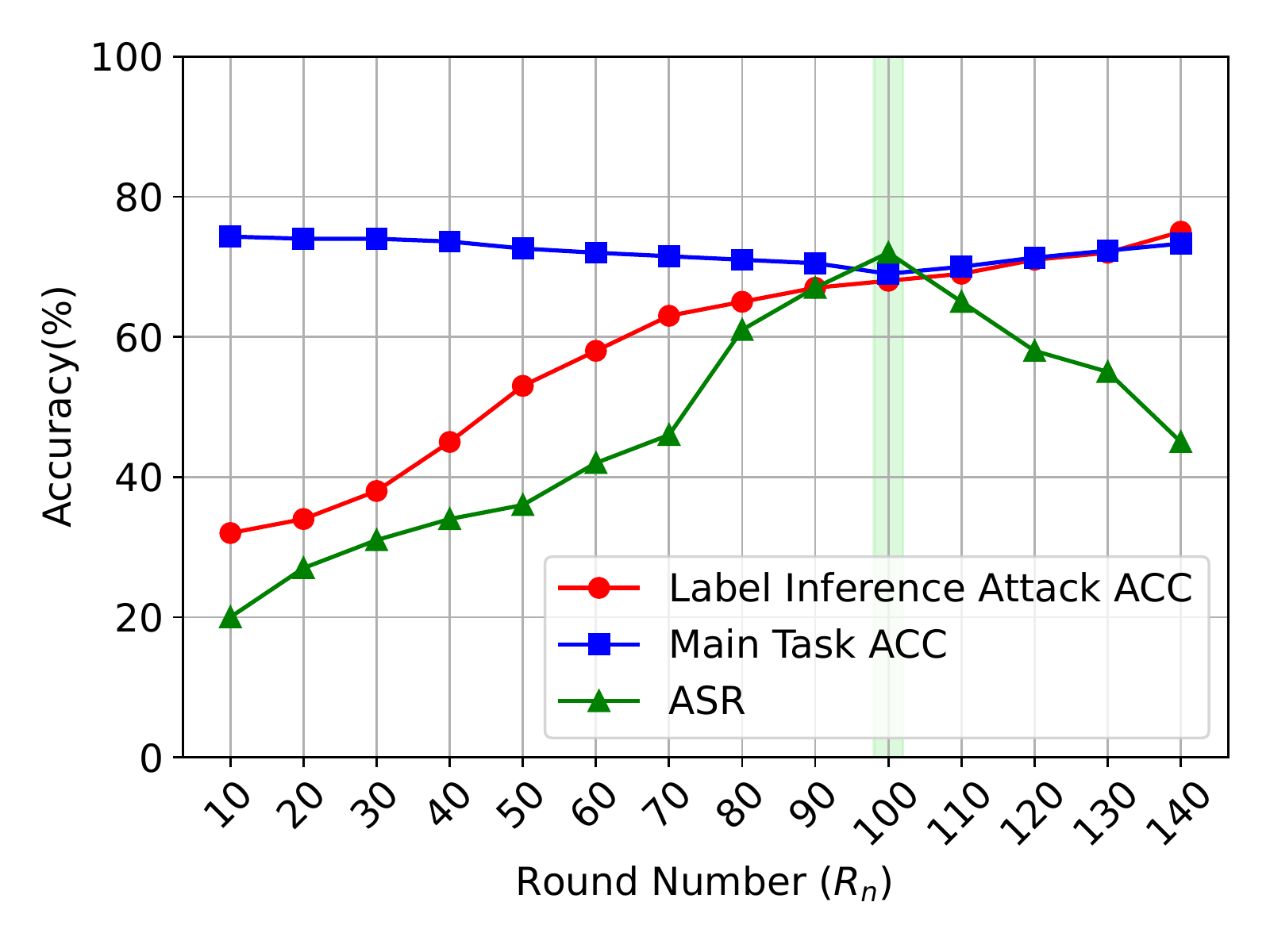}}
\subfloat[CINIC-10, 6 participants]{\label{CINIC-10-6party}\includegraphics[width=\tempwidth]{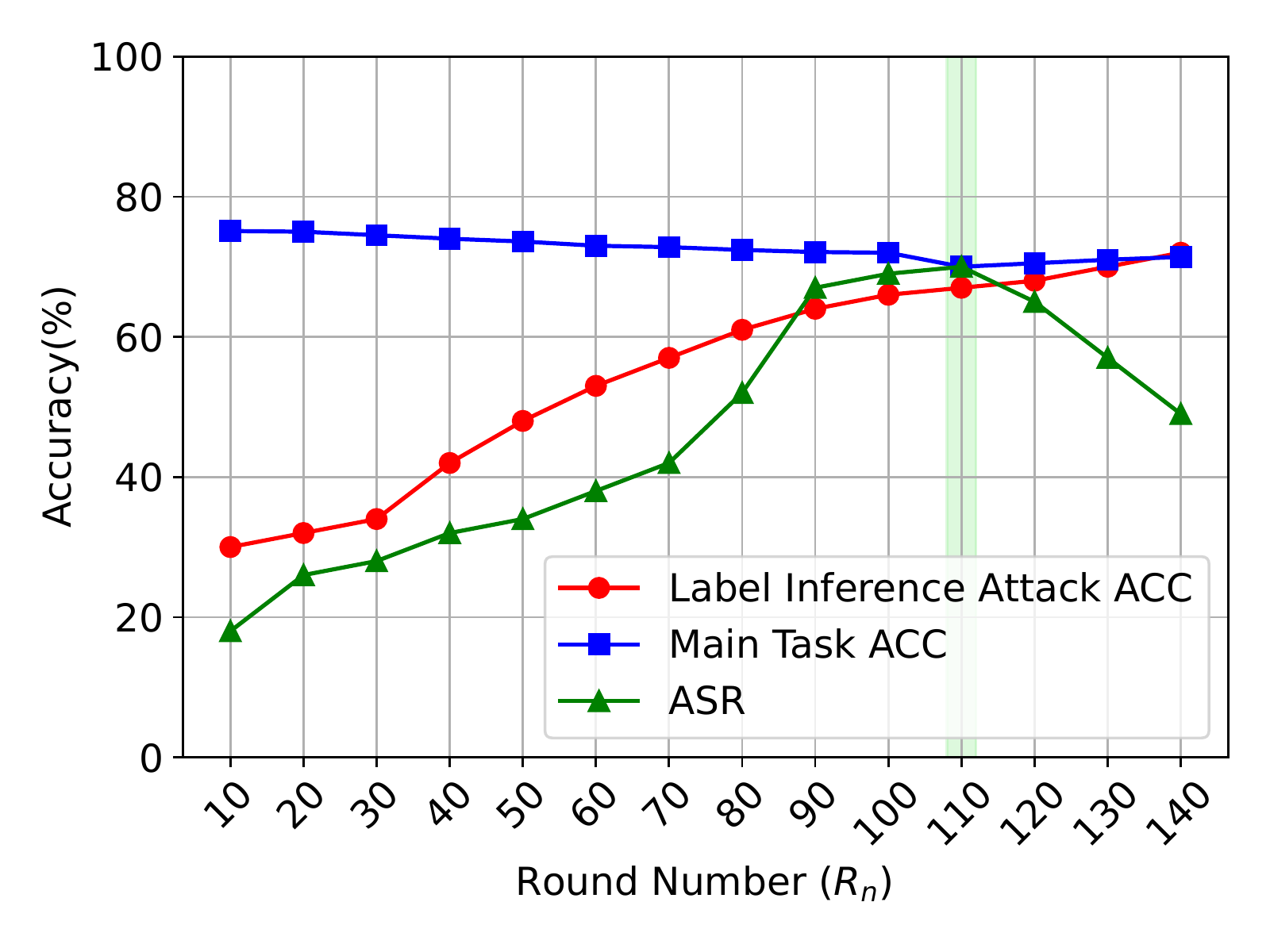}}
\subfloat[CINIC-10, 8 participants]{\label{CINIC-10-8party}\includegraphics[width=\tempwidth]{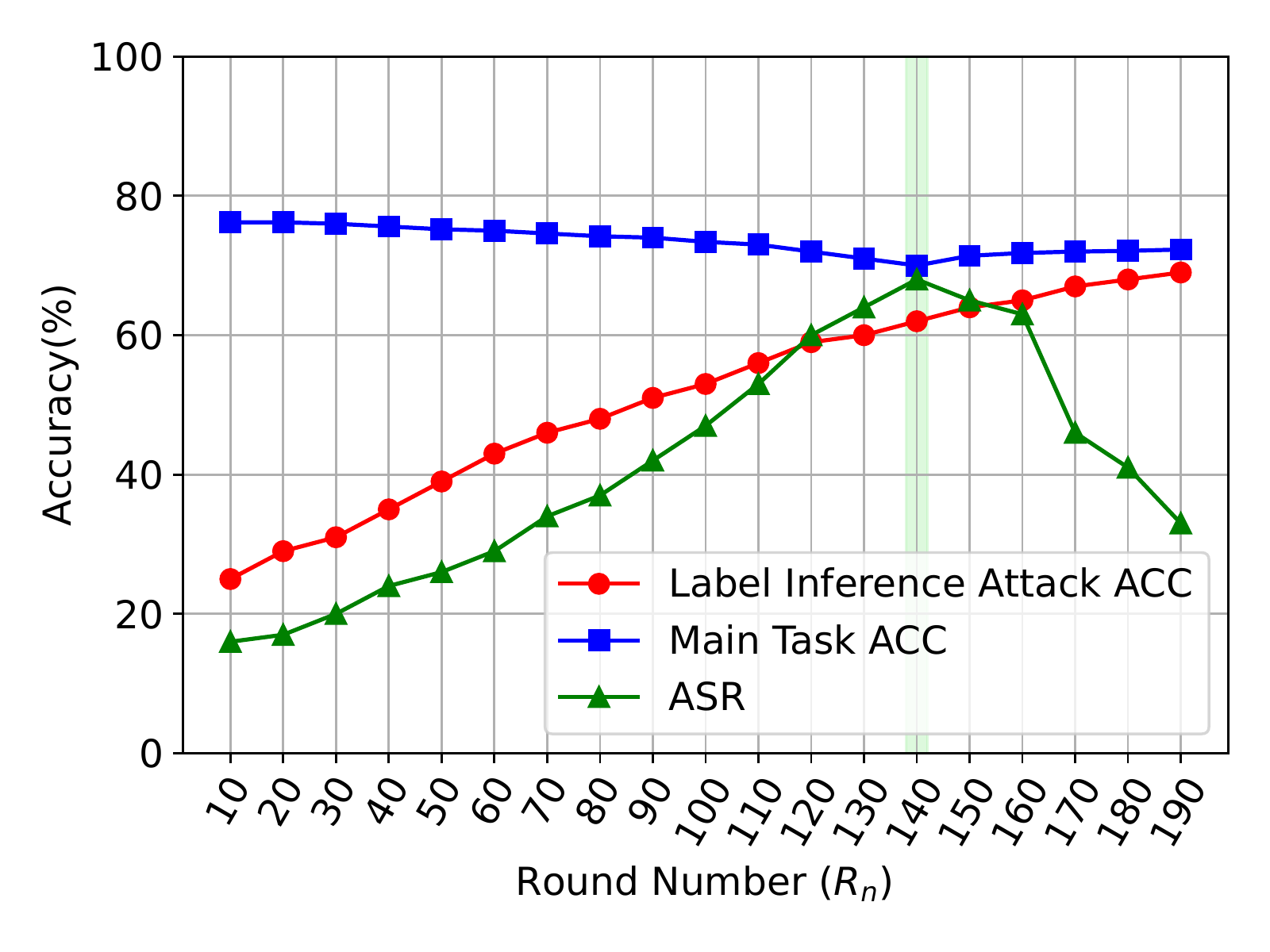}}
\subfloat[CINIC-10, 10 participants]{\label{CINIC-10-10party}\includegraphics[width=\tempwidth]{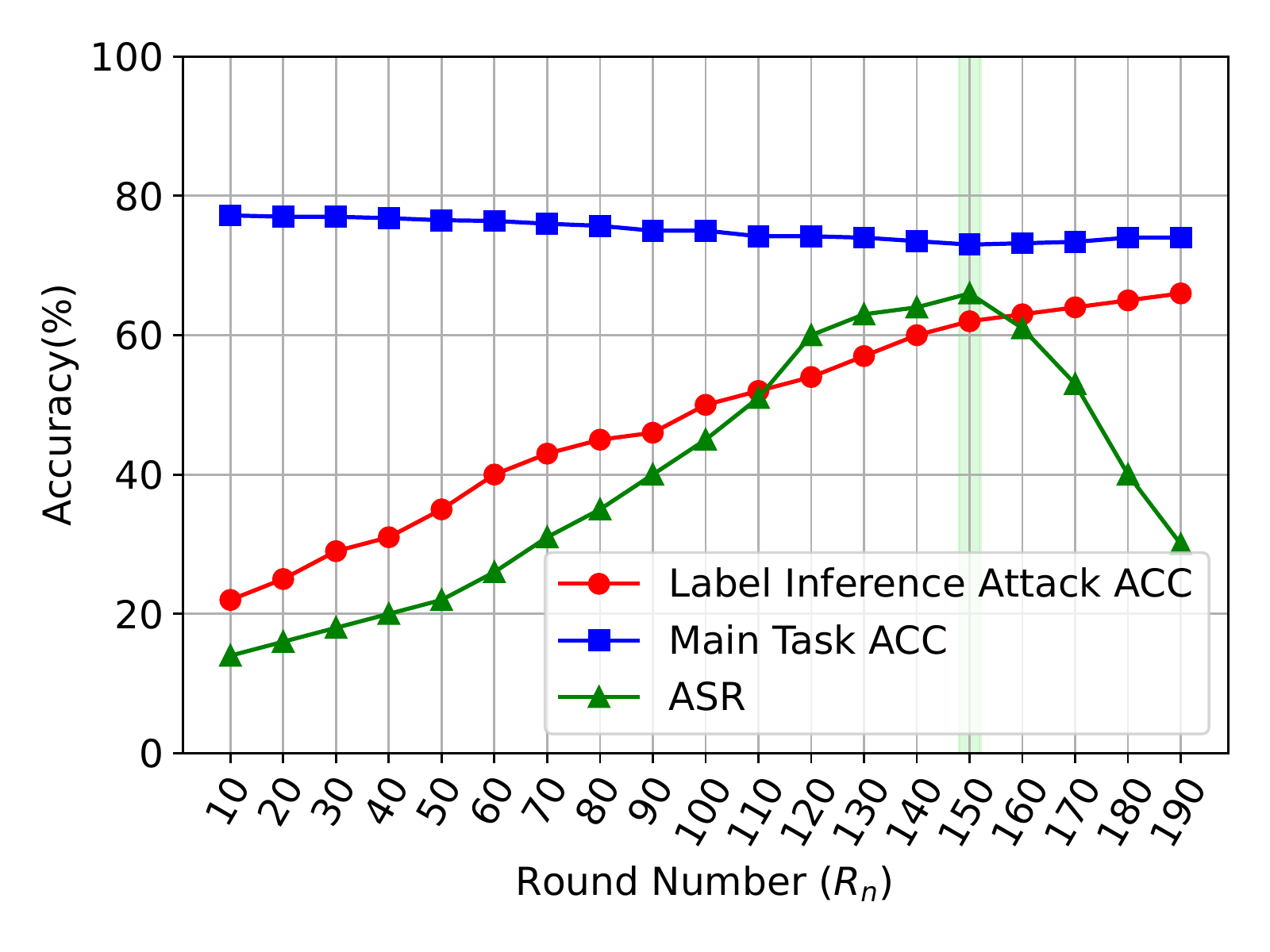}}\\
[-2ex]
\subfloat[Criteo, 4 participants]{\label{Criteo-4party}\includegraphics[width=\tempwidth]{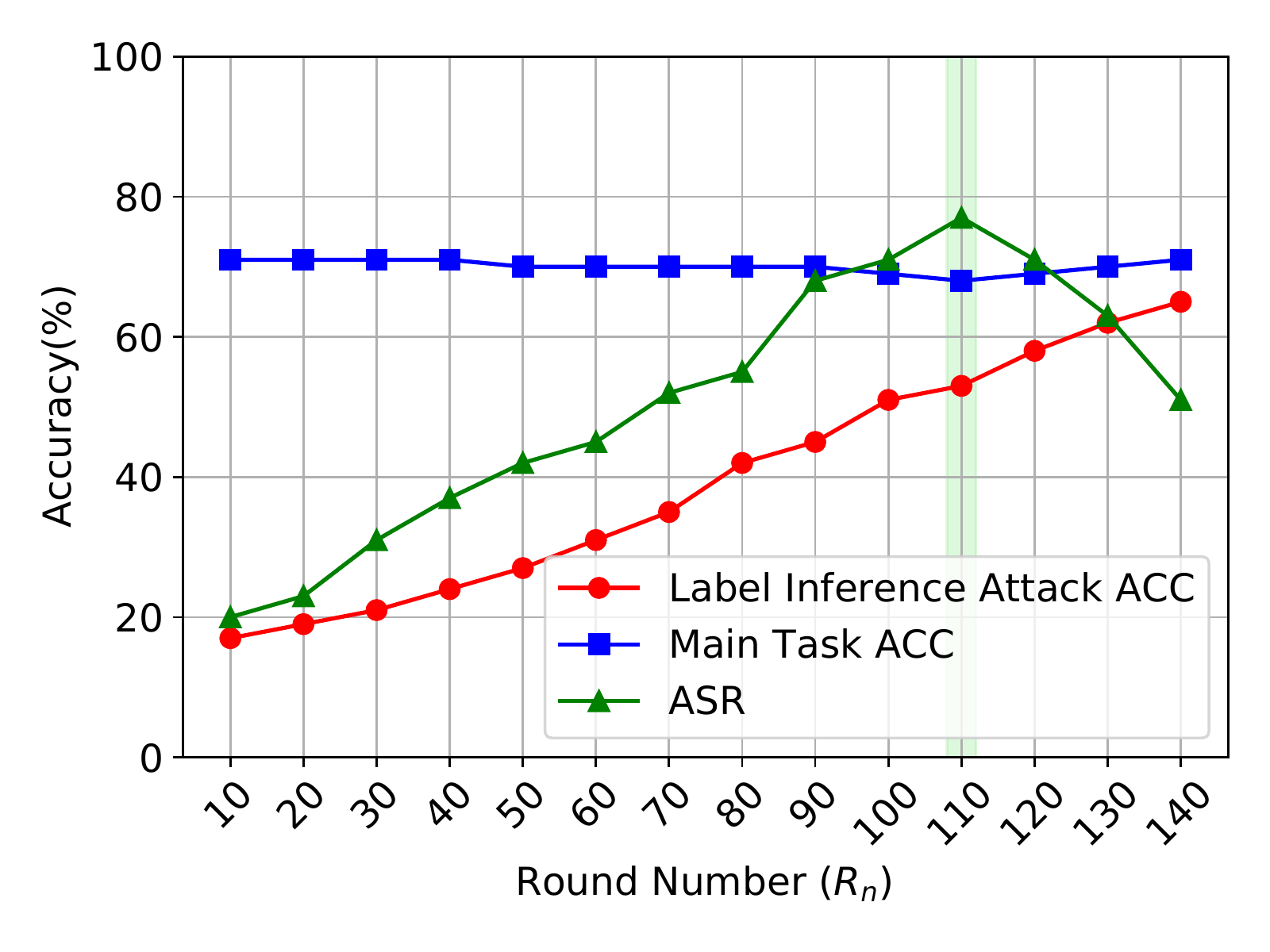}}
\subfloat[Criteo, 6 participants]{\label{Criteo-6party}\includegraphics[width=\tempwidth]{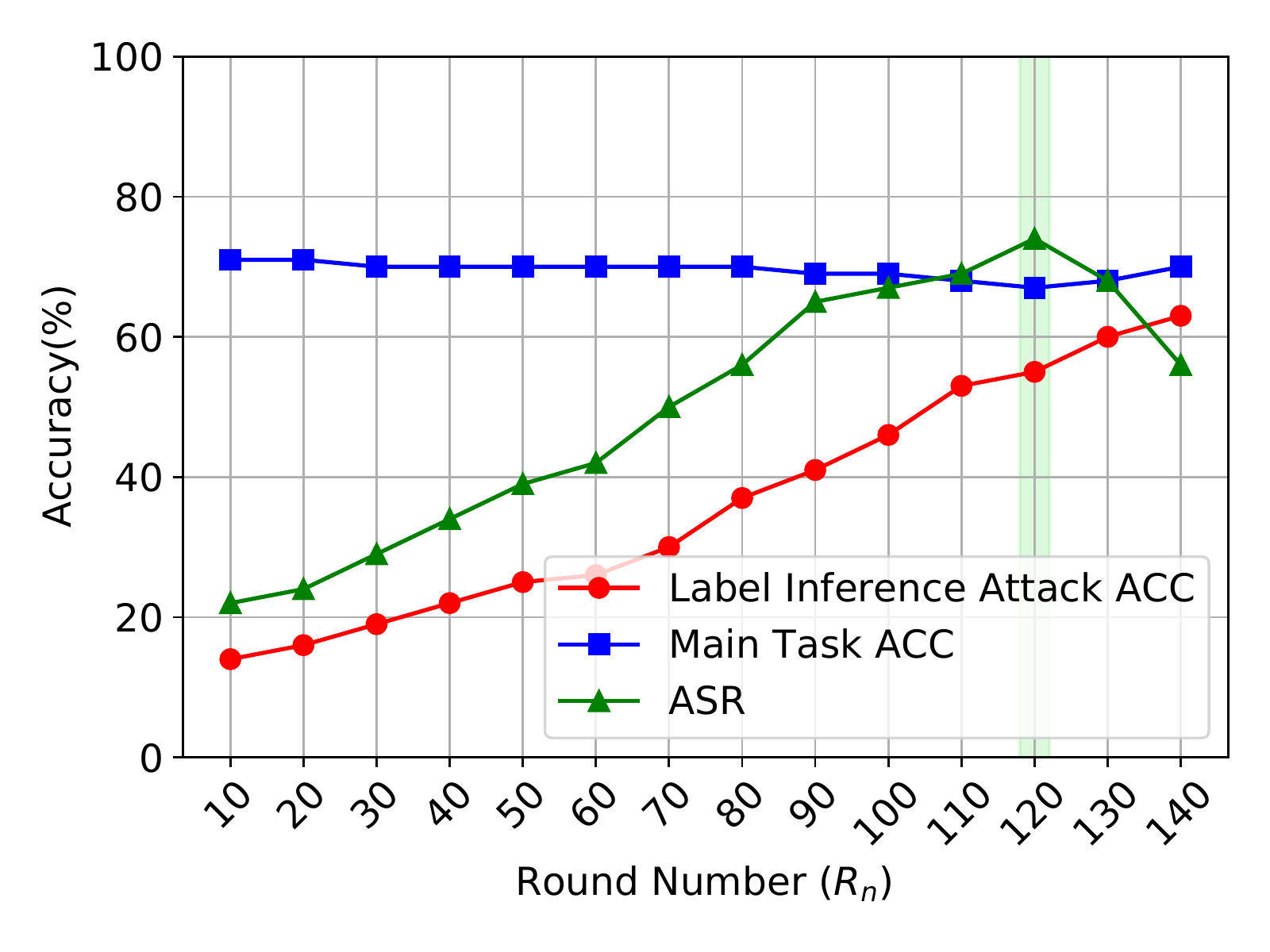}}
\subfloat[Criteo, 8 participants]{\label{Criteo-8party}\includegraphics[width=\tempwidth]{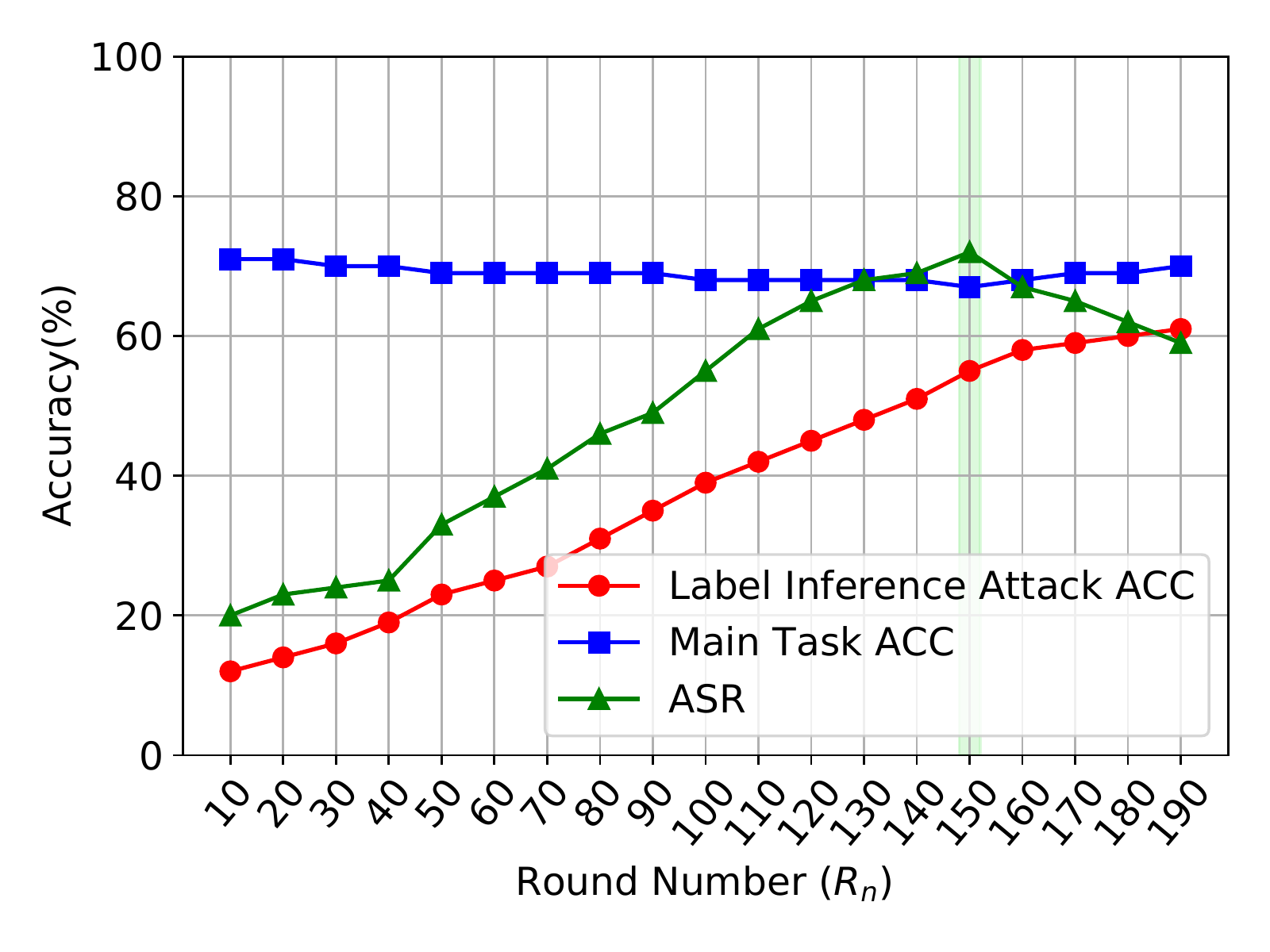}}
\subfloat[Criteo, 10 participants]{\label{Criteo-10party}\includegraphics[width=\tempwidth]{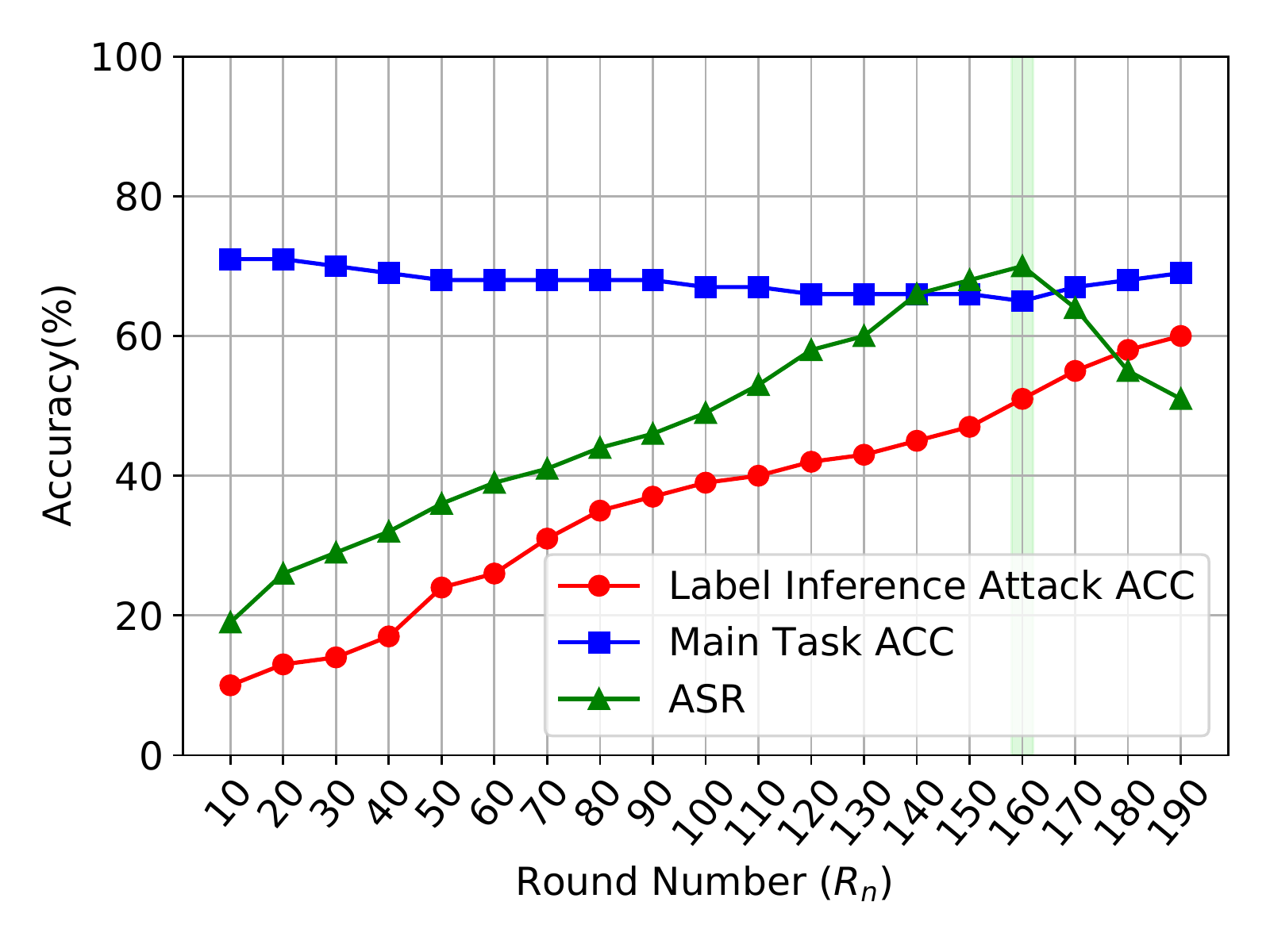}}
\caption{Main task accuracy (MTA), accuracy of the label inference attack, and ASR at each round.}
\label{fig:multipartybackdoorperformance}
\end{figure*}

\begin{table}[t]
\centering
\begin{tabular}{lrrrr}
\toprule
\textbf{Dataset}& \textbf{4-Party}& \textbf{6-Party}& \textbf{8-Party}& \textbf{10-Party}  \\  
\midrule
CIFAR-10 & 0.821   & 0.829   & 0.832  & 0.841     \\ 
CIFAR-100 & 0.749 & 0.755 & 0.774 & 0.780    \\ 
CINIC-10& 0.743 & 0.751 & 0.762  & 0.772	  \\ 
\bottomrule
\end{tabular}
\caption{Main task accuracy (MTA) in multi-party settings, under no attack.}
\label{tab:maintaskmultiparty}
\end{table}

As having more participants requires more rounds to make the VFL model converge, we increase the total number of rounds. 
In Fig.~\ref{fig:maintaskmultiparty}, we report the main task accuracy for an increasing number of training rounds. 
Based on the observations, we set the total number of rounds for the attack experiments to $150$ for 4 and 6 participants and to $200$ for 8 and 10 participants. 
Table~\ref{tab:maintaskmultiparty} reports the accuracy of the converged models for multiparty experiments under no attack. 
Perhaps unsurprisingly, more parties increase the accuracy of the main task as the dimensionality of the input feature embeddings to the top model is likely increased too. 
In other words, higher dimensional feature embeddings help enlarge the classification margin between different classes~\cite{Emami2020icml}, which improves classification accuracy.

As discussed in Section~\ref{twopartysection}, it is challenging to set the value of $R_n$. 
Thus, we measure the performance of the attack with different $R_n$ values for an increasing number of participants; see Fig.~\ref{fig:multipartybackdoorperformance}.
Overall, our backdoor attack is effective in a multi-party setting with a slight reduction in the accuracy of the main task.
For instance, on CIFAR-100, in a setting of 8 participants where one of them is the adversary, BadVFL's ASR reaches 0.74, while the main task accuracy decreases from 0.77 to 0.73. 
Additionally, increasing the value of $R_n$ leads to higher ASR when the number of participants also increases.
This is because the label inference attack achieves the desired accuracy for the backdoor attack in the later rounds when the number of participants increases. 
Compared to the two-party attack (see Figure~\ref{fig:ASRallthreedatasets} and Table~\ref{tab:backdoormaintaskinroundnumber}), the poisoning effects of the backdoor attack remain persistently even when the adversary controls only one of the 10 participants. 
This shows that having more participants does not reduce ASR, which poses a severe threat to model integrity in VFL. 

Our experiments assume a non-overlapping feature distribution among the participants. 
However, as pointed out by Fu et al.~\cite{fu2022label}, if an overlapping of features exists, the adversary can improve the label inference attack, which decreases $R_n$, and as a result, increases the attack performances of BadVFL. 
We leave this as part of future work. %

\subsubsection{Multi-Attacker mode}
Next, we experiment with a collective backdoor attack executed by multiple attackers. 
We first break down a trigger pattern into sub-trigger patterns and inject them separately into the data samples of each participant controlled by the adversary following the label inference and trigger injection stage.  
In the testing phase, we inject all the sub-triggers into the malicious participants in parallel to conduct the backdoor attack.  
We experiment with the CIFAR-10 dataset in a setting involving 10 participants, and 2 of them are malicious. 

\begin{figure}[t]
\centering
\includegraphics[width=0.35\textwidth]{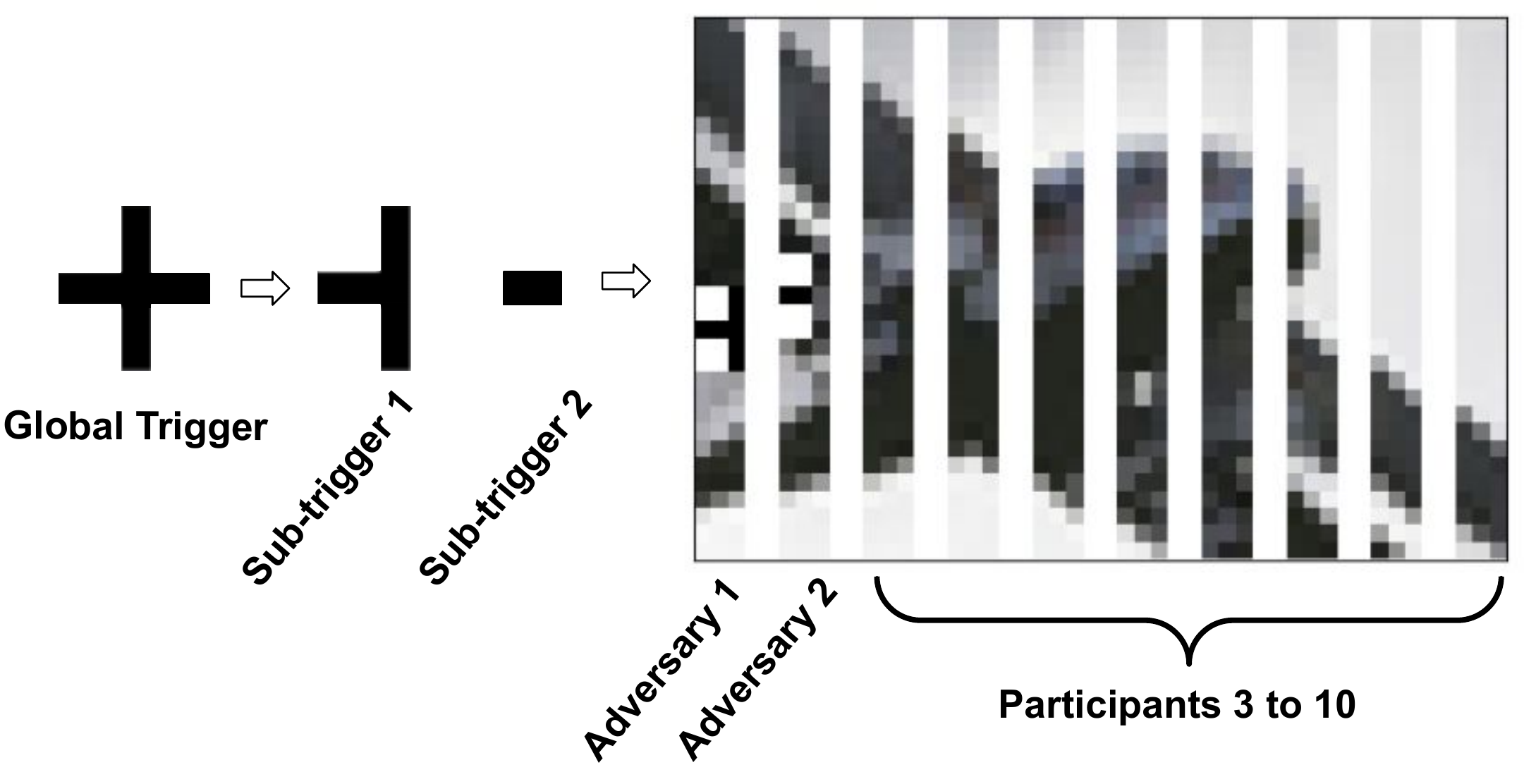}
\caption{Illustration of the distributed backdoor attack with 10 VFL participants in VFL and 2 attackers.}
\label{fig:DBAtwoattacker}
\end{figure}

As shown in Fig.~\ref{fig:DBAtwoattacker}, we split the window of size $5\times5$, marked with a cross sign pattern, vertically, into two triggers of sizes $5\times3$ and $5\times2$.
Each malicious participant uses them independently. 
The attack follows the same structure as discussed in Section~\ref{backdooralgorithmflow}. 
Both attackers share the same auxiliary dataset of size 40 for the label inference phase and use a poisoning budget of 20\%. 
We run the experiment for 200 rounds, with the backdoor attack initiated at $R_n=140$, which we set empirically by taking into account the single attacker mode experiments, as presented in Fig.~\ref{fig:multipartybackdoorperformance}.

\begin{table}[t]
\centering
\begin{tabular}{cc|rr|rr}
\toprule
\textbf{Global} & {\bf Sub} & \multicolumn{2}{c|}{\textbf{Multi-Attacker}}     & \multicolumn{2}{c}{\textbf{Single-Attacker}}    \\
{\bf Trigger}         &   {\bf Triggers}      & \multicolumn{1}{c}{\textbf{ASR}} & \textbf{MTA} & \multicolumn{1}{c}{\textbf{ASR}} & \textbf{MTA} \\ \midrule
5$\times$5 & 5$\times$3,  5$\times$2   & 0.782  &         0.803       & 0.771  &   0.775     \\        
\bottomrule
\end{tabular}
\caption{Main task accuracy (MTA) and ASR in Multi-Attacker vs.~Single-Attacker modes in a 10-party VFL setting on CIFAR-10.} 
\label{tab:twovsoneattacker}
\vspace{0.2cm}
\end{table}

Table~\ref{tab:twovsoneattacker} compares the attack performance with multiple malicious participants to that with only one participant. 
We include 10 participants in both cases and choose to inject the global trigger using the optimal selection strategy. 
The poisoning budget is 20\% in both cases. 
The only difference is the two attackers split the trigger, and each of them uses one of the sub-triggers. 
Whereas the single attacker mode uses the trigger directly in the attack. 

With a window size of $5\times5$ in the multi-attacker setting, BadVFL's ASR is slightly higher than that of the single-attacker setting (0.78 vs.~0.77). 
The main task accuracy also improves (0.80 vs.~0.78). 
We believe this is due to two reasons.
First, involving more malicious participants improves the flexibility of the attack. 
VFL resembles multi-view learning \cite{ZHAO201743} by design; the feature subset hosted by each participant can be considered to be one separate view of the training data set. 
Decomposing the global trigger to the malicious participants introduces the desired backdoor poisoning effects into multiple views of the training data. 
Thus, it enhances the association between the triggered input and the target class label, which yields higher ASR. 
Second, injecting the sub-triggers instead of the global trigger into each malicious participant reduces the negative impact on the main learning task over each participant. 
Intuitively, smaller sub-triggers introduce less perturbation to the training data owned by each malicious participant, which helps preserve better accuracy.

\subsection{Take-Aways}
Overall, we find that BadVFL provides an effective backdoor attack with minimal impact on the performance of the primary task. 
There are different factors that need to be taken into account when performing the attack.
First, BadVFL consists of two phases: label inference and backdoor attack. 
The main settings that affect the former include the size of the auxiliary dataset and $R_n$. 
Our experiments show that finding an optimal $R_n$ greatly impacts the performance of both phases. 
The backdoor phase of BadVFL, on the one hand, requires labels to be inferred to a certain extent and, on the other hand, necessitates some rounds to inject the trigger.
We find that an accuracy of approximately 60\% for the label inference phase is necessary for a successful backdoor attack; however, we still need the second phase of the attack to have sufficient rounds to inject the trigger.

Second, the number of participants in the VFL system also has a non-negligible impact on the attack performance. 
With the same number of attackers, the effectiveness of the attack decreases with an increasing number of participants. %
In addition, with more participants, the value of $R_n$ needs to be increased to leave enough rounds for the label inference phase. 
Moreover, we can execute the BadVFL attack with multiple adversaries where the trigger is divided into sub-triggers, and the performance of the attack is slightly better than with a single adversary, but the main task accuracy is much higher.

Finally, the way of selecting the source/target classes significantly affects the attack performance. 
Their optimal selection, in comparison to selecting at random, always results in a better attack, with only a minor decrease in the main task accuracy. 
Additional factors like trigger window size and poisoning budgets also impact the performance of BadVFL: bigger triggers and larger poisoning budgets introduce greater poisoning efforts into the trained classifier, which ultimately increases the attack performance.

\section{Countermeasures}\label{sec:countermeasure}
In this section, we discuss three possible countermeasures to mitigate the BadVFL attack, using: 1) Neural Cleanse~\cite{wang2019neural}, 2) Differential Privacy noise, and 3) Anomaly Detection based defenses.

\descr{Neural Cleanse (NC) on the VFL Server.}
NC~\cite{wang2019neural} is a technique geared to identify and mitigate backdoor attacks in neural networks. 
More precisely, it first identifies candidate neurons that may be involved in a backdoor attack by analyzing the network's behavior on different inputs.
Then, it uses a pruning algorithm to remove these neurons from the network. 
Finally, it retrains the network to ensure that the performance is not impacted. %

NC should not be used during training but at inference time. 
In order to use NC in VFL, one possible strategy could be to have the server periodically evaluate the global model for the presence of a backdoor attack by testing the model on a validation dataset that includes the trigger pattern used in the backdoor attack. 
If the model's predictions on the validation dataset are significantly different from the expected output, this could indicate the presence of a backdoor attack.
However, since in our proposed attack, the server takes feature embeddings as input and not the triggered images, then NC cannot defend against the attack.

\descr{Differential Privacy.}
Another defense mechanism against backdoor attacks is to apply differential privacy~\cite{abadi2016deep} (DP) noise to the input embeddings transmitted from participants to the top model.
\edits{In fact, DP has been used in HFL systems as a defense mechanism against potential backdoor and distributed data poisoning attacks \cite{Nguyen22Flame, Xie21ICML, Naseri22ndss}.
Note that this approach is different from the typical use of DP to establish privacy protections for training data or models. 
Rather, one uses the random noise perturbation produced by a DP mechanism as a backdoor defense method.  
The key idea in  \cite{Nguyen22Flame, Xie21ICML, Naseri22ndss} is to introduce additional randomized and learning-objective independent perturbation to the global/local model to prevent the model from learning the association between the trigger-embedded input and the attacker-desired class labels. 
As discussed in \cite{Xie21ICML, Nguyen22Flame}, a $(\epsilon,\delta)$-differentially private global model is also backdoor-free by adding Gaussian random noise of a certain level of variance (see Theorem 2 in \cite{Xie21ICML} and Theorem 1 in \cite{Nguyen22Flame}).}

\begin{figure}[t]
\centering
\includegraphics[width=0.35\textwidth]{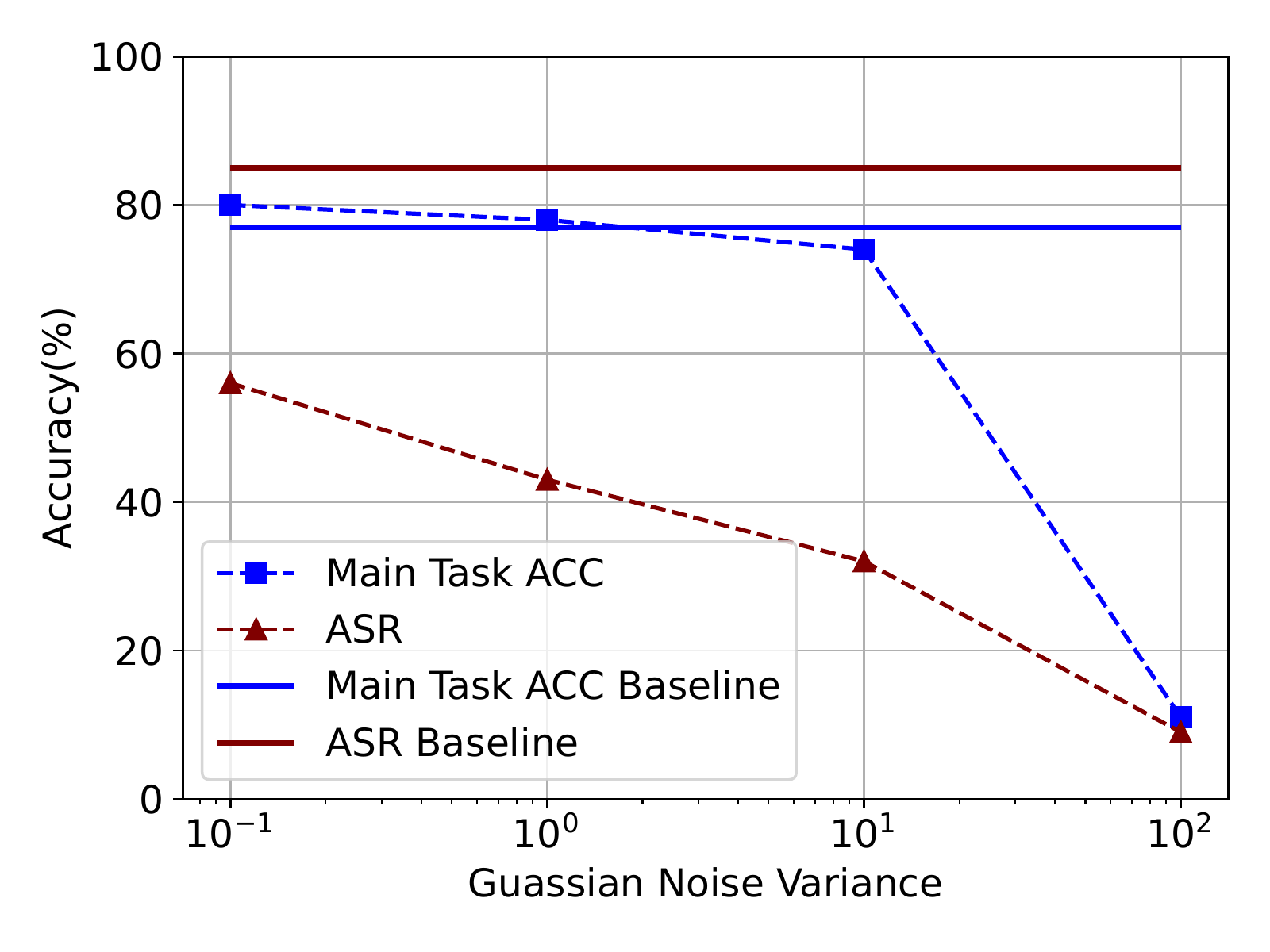}
\caption{Adding differential privacy noise as a defense in the two-party VFL setting on CIFAR-10.}
\label{fig:DPnoise}
\end{figure}

We experiment with a two-party setting on the CIFAR-10 dataset, where one participant performs the backdoor attack.
We use the trigger signal of the size $5\times5$ and apply the poisoning budget of 10\%, following the setting in Table~\ref{tab:backdoormaintaskinroundnumber}. 
To defend against the attack, the server adds Gaussian noise with an increasingly larger variance from $1e-1$ to $1e2$ to the feature embeddings submitted by each participant to the server in each round. %
We report the average attack performance, over five independent runs of the BadVFL attack, in Fig.~\ref{fig:DPnoise}.
ASR drops from $85\%$ to, respectively, $58\%$ and $42\%$ with the variance of the DP noise as $1e-1$ and $1$, respectively.
There is almost no reduction in utility.
While in HFL, differential privacy-based defenses usually result in significant losses in the main task accuracy, here we see that the feature embedding in VFL shows much stronger resilience to the DP noise.
We believe this is due to the feature embeddings being less sensitive than the raw input or gradients to the FL system, as discussed in recent work on federated feature learning~\cite{shen2023share}. 
Although additional work is needed to explore this further, we are confident that DP-based defense mechanisms can potentially achieve reasonable utility-integrity trade-offs in the context of backdoor attacks against VFL.

\descr{Anomaly Detection.}
Finally, we consider performing anomaly detection over the feature embeddings of each class.
In our backdoor attack, the attacker replaces some of the training data points belonging to the target class.
Thus, the server could perform anomaly detection on the feature embeddings of the same class submitted by each participant.

For example, for a given class $k$ and a participant $i$, $f_{i,k}$ are the embeddings of training data of the class $k$ submitted by the participant $i$.
The server can use \textit{Isolation Forest}~\cite{liu2008isolation} on $f_{i,k}$ to identify the $p\%$ training data points of the class $k$ with the highest anomaly scores.
The server then excludes these training data points from computing the gradients of the top layer model and updating the model.
We can simply set $p\%$ (the excluded data points) to be the same as the number of the poisoned training data points, known as the poisoning budget.

\begin{table}[t]
\centering
\begin{tabular}{l|rr|rr}
\toprule
& \multicolumn{2}{c|}{\textbf{No Defense}}     & \multicolumn{2}{c}{\textbf{Anomaly Detection}}    \\
{\bf Dataset}                  & \multicolumn{1}{c}{\textbf{ASR}} & \textbf{MTA} & \multicolumn{1}{c}{\textbf{ASR}} & \textbf{MTA} \\ \midrule
CIFAR-10          & 0.85  &         0.77       & 0.34  &   0.75             \\ 
\bottomrule
\end{tabular}
\caption{Main task accuracy (MTA) and ASR when anomaly detection is applied against BadVFL.} 
\label{tab:anomalydetection}
\vspace{0.2cm}
\end{table}

We explore this idea experimentally, conducting an anomaly detection experiment in a two-party setting over CIFAR-10, similar to the one depicted in Fig~\ref{3metrics100rounds_2party_cifar10}, with the presence of a single adversary.
The value of $R_n$ is set to $60$.
We provide the server with the advantage of knowledge regarding the poisoning budget value.
Note that it is just the value, not the poisoned data points. 

The results of our experiments are presented in Table~\ref{tab:anomalydetection}. 
We find that anomaly detection is overall effective, with a minor decrease in the main task accuracy.
However, computing the inter-class distance in the feature embedding space results in a large overhead for the server, as the computational cost increases quickly with an increasing number of training data instances. 
Therefore, this is not a viable defense method when the server has limited computation resources.

\section{Related Work}\label{sec:related}
\noindent\textbf{Backdoor attack in HFL.} Early studies on backdoor attacks against HFL \cite{howtobackdoor,foolsgold} assume that each malicious participant independently trains a local model without collusion between them. 
The resulting poisoned local models tend to share similar parameter values and deviate significantly from the local models submitted by benign participants. 
These attacks can thus be mitigated by Byzantine-robust aggregation methods and defense methods against sybil attacks like \textit{Foolsgold}~\cite{foolsgold}.

More advanced distributed backdoor attacks \cite{alittle,sun2019really} consider how to avoid being flagged by Byzantine-robust aggregation rules~\cite{mult-krum,bulyan,trimmed_mean}. 
They clip the parameters of poisoned local models according to the bounds on the parameter values of benign local models. 
These methods either assume the parameter values of benign local models are IID Gaussian variables so that the variance bounds of poison-free parameter values can be estimated~\cite{alittle}, or assume that the poison-free parameter bounds are known as prior knowledge~\cite{sun2019really}. 
These assumptions very rarely hold in practice, especially in non-IID FL scenarios. 
As a result, manually configured thresholds for parameter clipping may be overestimated (downgrading the learning capability) or underestimated (failing the attack task).

\textit{DBA}~\cite{DBA} manages backdoor attacks by manually decomposing global triggers into separate local triggers and assigning separate local triggers to each malicious participant. 
Malicious participants learn to fit different local triggers and thus have dissimilar poisoned local models to bypass \textit{Foolsgold}~\cite{foolsgold}. 
However, manually decomposed triggers may cause unexpectedly large deviations of poisoned local models from the benign ones.
Therefore, this method fails to attack the Byzantine-robust aggregation methods such as \textit{Krum}~\cite{mult-krum} and \textit{Bulyan}~\cite{bulyan}. 

\descr{HFL vs.~VFL.} While previous work has more extensively studied backdoor attacks in Horizontal Federated Learning (HFL), little work has focused on the feasibility of these attacks in Vertical Federated Learning (VFL). 
Compared to HFL, there are two main challenges in building backdoor attacks in VFL.
First, the malicious participants in VFL cannot modify the class labels, which prevents the attacker from generating and introducing directly the backdoor training samples annotated with the attack's desired class label. 
Second, the attacker can only manipulate the features hosted by the malicious participants, which includes only a fraction of the training features, thus potentially weakening the strength of the poisoning efforts. 

\descr{Threats against VFL.} 
Most previous research on VFL security \cite{fu2022label,VFL_CAFE, VFL_Unleashing} focuses on privacy leakage. 
Fu et al.~\cite{fu2022label} investigate a new attack launched by a local participant who tries to infer the training data's labels by fine-tuning their bottom model with some auxiliary labeled data. 
By maliciously increasing the learning rate of the local optimizer, the adversary can make the VFL model heavily dependent on the bottom model of the malicious participant, thus further improving their ability to infer labels. 
Also, Pasquini et al.~\cite{VFL_Unleashing} demonstrate that a malicious server could infer local participants' training data by actively \textit{hijacking the VFL learning process}.
They also perform a property inference attack based on similar hijacking; the drawback is that it totally replaces the main learning task and thus is unclear whether is applicable in reality. 

By contrast, there is very little work studying backdoor/data poisoning attacks in VFL. 
Inspired by the clean-label backdoor attacks adopted in centralized learning scenarios, BadVFL demonstrates for the first time the feasibility of mounting backdoor attacks against VFL; in the process, we discuss what factors impact their performance and the viability of mitigation strategies.
Overall, we believe our work will pave the way for further research in this field, both in terms of attacks and defenses.

\section{Discussion \& Conclusion}\label{sec:conclusion}

\subsection{Summary}
This paper presented a novel clean-label backdoor attack in VFL, named BadVFL.
First, BadVFL tunes the feature embeddings of the poisoned samples belonging to a target class; the goal is to push the slightly perturbed data of the target class toward the trigger-embedded data of a source class in the feature embedding space. 
The feature-level poisoning misleads the classification module in the top model and makes the classifier misclassify the trigger-embedded data of the source class from the perturbed data of the target class. 
Second, BadVFL computes the saliency map of training data to locate the trigger in the training data, which we show yields more effective attacks than randomly inserting the trigger.
 
Our results demonstrate the effectiveness of BadVFL with two or multiple participants in a target VFL system. 
Our experimental evaluation illustrates the persistent backdoor threat raised by BadVFL with varying trigger sizes, the number of attackers among the participants, and different strategies to select the source and target classes of the attack.
We also explore potential countermeasures against this attack, reviewing existing defenses against backdoor attacks in centralized learning/HFL and evaluating their effectiveness in mitigating BadVFL.
Our findings suggest that while techniques like DP and anomaly detection can possibly provide meaningful countermeasures against BadVFL in certain scenarios, there is still a need for further work to design robust, comprehensive, and scalable defenses.

\subsection{Future Work}\label{sec:futurework}

\descr{Dependency to Label Inference.}
Naturally, our work is not without limitations. 
One is that the adversary needs labeled auxiliary data to select the source/target classes. 
Part of our backdoor attack uses it to infer the labels. 
Although access to auxiliary data is a common assumption in literature, we empirically show we only need a handful of auxiliary data instances -- as little as 1\% of the training instances for the three image datasets we experiment with. %
In future work, we will investigate how to extend BadVFL to remove the dependency on the label inference phase and work in an auxiliary data-free way. 
For example, we could perform clustering of feature embeddings submitted by the adversary to estimate labels. 
Moreover, we will ground the association between the accuracy of label estimation and the attack performance on theoretical analysis. 

\descr{Impacting Factors.} We experimentally evaluated different factors impacting the attack accuracy (ASR) and that of the main task. 
For instance, deciding about $R_n$ has a crucial effect. %
A potential item for future work could involve novel methodologies to incorporate various factors, such as auxiliary dataset, poisoning budget, and window size, into the decision-making process for selecting the value of $R_n$, as this decision must be made by the attacker during training.

\descr{Defenses.}
Despite our successful implementation of effective defenses against the backdoor attack, additional research could explore additional defenses in the VFL setting, particularly with more participants and attackers.
Moreover, the anomaly detection defense requires the server to know the poisoning budget -- something we plan to investigate further.

\descr{Feature Correlation.}
It is possible that the feature subsets hosted by different participants are intrinsically correlated, which may boost the attack. 
We will discuss how to organize effective attack collusion between correlated feature subsets to 
inject the trigger together.

\descr{Acknowledgments.} This work was partially supported by a Microsoft Research PhD/EPSRC CASE Studentship on ``Privacy in Distributed and Collaborative Learning'' as well as an Amazon Research Award on ``Studying and Mitigating Inference Attacks on Collaborative Federated Learning.''

{\small
\bibliographystyle{abbrv}

}

\end{document}